\documentclass{article}

\PassOptionsToPackage{numbers, compress}{natbib}

 \usepackage[preprint]{neurips_2026}

\usepackage[utf8]{inputenc} 
\usepackage[T1]{fontenc}    
\usepackage{hyperref}       
\usepackage{url}            
\usepackage{booktabs}       
\usepackage{amsfonts}       
\usepackage{nicefrac}       
\usepackage{microtype}      
\usepackage{xcolor}         
\usepackage{xspace}
\usepackage{subcaption, colortbl}
\usepackage{titletoc}       

\usepackage{amsmath,amsfonts,bm}

\def\eqref#1{equation~\ref{#1}}

\def\1{\bm{1}}

\DeclareMathAlphabet{\mathsfit}{\encodingdefault}{\sfdefault}{m}{sl}
\SetMathAlphabet{\mathsfit}{bold}{\encodingdefault}{\sfdefault}{bx}{n}

\usepackage{makecell, graphicx, romannum, amssymb, amsmath, ccaption, setspace, algorithm, algpseudocode, enumitem, pifont, etoolbox, dsfont, fontawesome5, longtable, array, arydshln, bbding, multicol, multirow}

\newfixedcaption{\outfigcaption}{figure}
\newfixedcaption{\outtabcaption}{table}

\usepackage{tcolorbox,titletoc}
\newcommand{\authcount}[1]{}
\usepackage{xstring,catchfile}

\usepackage{hyperref}
\usepackage{cleveref}
\usepackage{tabularx}

\newcommand{\method}{\textsc{ST-DiffEye}\xspace}
\title{\method: Diffusion-based Continuous Gaze Generation via Joint Scanpath-Trajectory Modeling}

\author{%
  Brian Nlong Zhao
  \qquad
  Ozgur Kara\setcounter{footnote}{1}\thanks{Corresponding author}
  \qquad
  Junho Kim
  \qquad
  James M. Rehg
  \\\\
  University of Illinois Urbana-Champaign
}
\begin{document}
\pagenumbering{arabic}
\maketitle

\begin{abstract}
\label{sec:abstract}

We study the problem of human gaze modeling, which aims to generate the gaze patterns a viewer produces while observing a visual stimulus. Gaze is primarily captured through two modalities: continuous eye-tracking trajectories, which describe fine-grained motion dynamics, and discrete scanpaths, which describe high-level fixation structure. Because gaze varies substantially across viewers and trials, we treat this variability as a defining property rather than noise and model gaze as a stochastic generative process. Existing generative gaze models supervise on only one of these two representations in isolation. We hypothesize that trajectories and scanpaths describe gaze at complementary scales and are jointly informative during training, and test this hypothesis through \method, a joint trajectory-scanpath diffusion framework that couples both modalities by concatenating them as an additional raw input channel, requiring no architectural overhead beyond an input and output channel expansion. We further introduce a principled evaluation framework based on the Continuous Ranked Probability Score (CRPS), which generalizes any existing sequence similarity metric into a proper scoring rule that jointly assesses the accuracy and diversity of generated gaze. Experiments on task-driven visual search, covering both target-present and target-absent scenarios, and on free-viewing benchmarks demonstrate state-of-the-art performance. These results, along with detailed ablations, confirm the benefit of joint modeling and the value of distribution-aware evaluation in capturing the intrinsic variability of human gaze. Project webpage: https://st-diffeye.github.io/

\end{abstract}
\section{Introduction}
\label{sec:introduction}

Human visual attention can be measured via eye tracking, which records a viewer's gaze position over time to produce a continuous trajectory of high-frequency $(x,y)$ coordinates \cite{yarbus2013eye, rothkopf2007task}. In the standard analysis pipeline, the continuous eye tracking trajectories are converted into a discrete representation called a scanpath, by identifying the fixations and the saccades that connect them \cite{Noton1971scanpath,NOTON1971929scanpath}. The scanpath representation is used in a broad set of application domains, ranging from cognitive psychology and advertising to virtual reality \cite{Moore2018application,10.1007/s10055-022-00738-zapplication}. 

With the emergence of generative AI, it is natural to ask whether realistic gaze behavior can be synthesized by a generative model that implicitly captures the dynamics of human attention. This is a challenging task because gaze emerges from a complex interaction between low-level image features, top-down goals, and individual traits, leading to high variability across subjects and even repeated viewings \cite{Torralba2006,doi:10.1073/pnas.1820553116}. Prior synthesis works have focused on two task conditions: 1) \textit{free-viewing}, in which the observer explores the stimulus image without any explicit goal; and 2) \textit{visual search}, in which the observer actively scans the stimulus to locate a designated target object category. Prior works have adopted the scanpath representation for three reasons. First, the scanpath provides a low-dimensional output representation whose length is bounded by the expected number of fixations. Second, a variety of widely-used metrics exist for comparing real and synthesized scanpaths.  Third, scanpaths are the predominant data modality in public eye tracking datasets. As a consequence, scanpath generation has emerged as the canonical task in gaze synthesis, and an on-going line of work has built increasingly expressive scanpath generators, from recurrent and reinforcement-learning formulations to recent diffusion-based models, in both free-viewing and visual search regimes \cite{scandiff,damico2024tppgaze,kummerer2022deepgaze}.

However, recent works~\cite{kara2025diffeye,jiao2026diffgaze} have departed from this convention by training generative models directly on raw eye-tracking trajectories. In particular, Kara et. al.~\cite{kara2025diffeye} demonstrated that training a trajectory generator to produce continuous gaze trajectory outputs, and then postprocessing the generated trajectories into scanpaths, yielded better performance in comparison to direct scanpath generation. They argued that continuous trajectories preserve information, for example saccade kinematics and gaze dynamics during fixation, which is useful for generative modeling and absent from the scanpath representation.

\begin{figure*}[t]
    \centering
    \includegraphics[width=\linewidth]{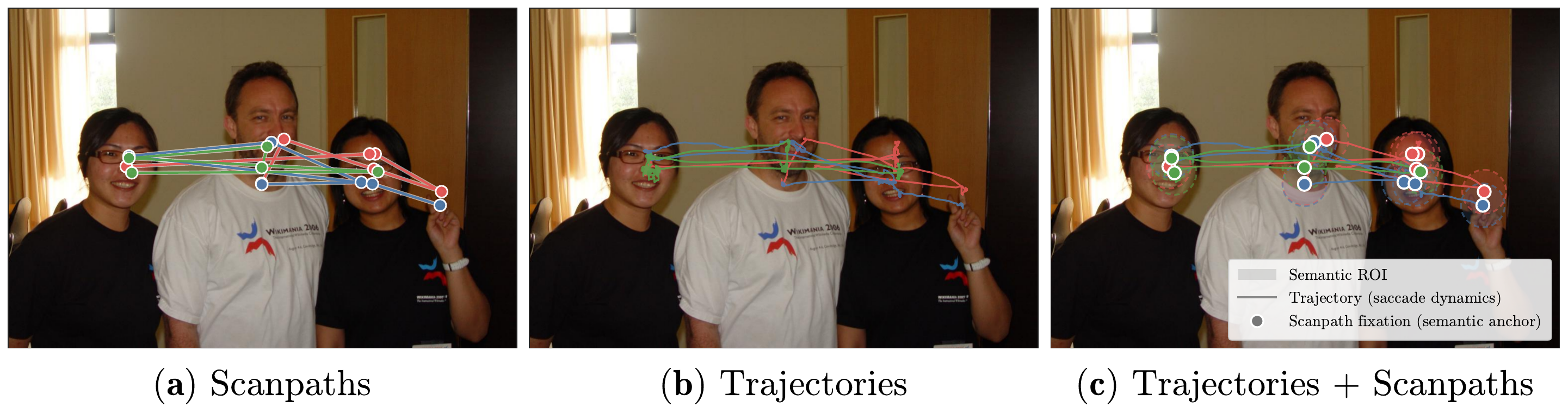}
    \caption{\textbf{Complementary representations of human gaze.} Prior work models gaze using either scanpaths for fixation-level structure (a) or continuous trajectories for fine-grained motion dynamics (b). Our approach jointly leverages both (c), aligning fixation anchors with the underlying trajectory along a shared time axis.
    }
    \label{fig:teaser}
    \vspace{-2em}
\end{figure*}

This paper advances the idea that continuous gaze trajectories and their associated scanpaths provide \emph{complementary} sources of information for gaze synthesis: Scanpaths provide a coarse structural summary of \emph{where} attention dwells, while eye tracking trajectories provide a fine-grained signal for \emph{how} the eye moves between fixations (Figure~\ref{fig:teaser}). This decomposition is reminiscent of prior work on multi-scale representations in computer vision~\cite{lin2017feature,wang2020deep}. We hypothesize that a generative model trained jointly on scanpaths and trajectories can exploit their complementary cues and produce samples that are both structurally coherent at the fixation level and dynamically variable at the trajectory level.


To test this hypothesis, we represent each scanpath as a continuous fixation index sequence of the same temporal length as its corresponding trajectory, and append it as additional channels to the trajectory data so that the diffusion model jointly denoises both modalities from a single augmented input. This formulation requires no architectural changes beyond a channel expansion in the data input and output, and induces a shared latent geometry in which the fixation structure of scanpaths and the continuous dynamics of trajectories are aligned along a common time axis. We show that the resulting joint representation improves both scanpath- and trajectory-level performance across free-viewing and task-driven benchmarks.

In addition, we observe that the standard sequence-level metrics compare generated scanpaths against human references using mean or best-match scores~\cite{jiao2026diffgaze,kara2025diffeye}, which reward point-estimate accuracy but do not capture distributional fidelity. KL-divergence-based protocols~\cite{damico2024tppgaze,scandiff} compare distributions of metric values, but provide only an indirect assessment that depends on the chosen metric. As a result, a model that repeatedly generates a small set of high-quality scanpaths may appear competitive with one that captures a diverse, human-like gaze distribution. To address this gap, we introduce a family of evaluation metrics built on the Continuous Ranked Probability Score (CRPS)~\cite{Gneiting01032007crps,mammadov2026variationalcrps}, a proper scoring rule that converts any existing distance- or score-based sequence metric into a distributional metric that jointly penalizes inaccuracy and insufficient diversity in a single interpretable number. We evaluate \method on COCO-FreeView~\cite{chen2021predictingcocofreeview}, MIT1003~\cite{Judd_2009mit1003}, and the task-driven COCO-Search18~\cite{chen2021cocosearch18} benchmarks, and report consistent gains over both trajectory-only and scanpath-only baselines. Our contributions are summarized as follows:

\begin{itemize}[itemsep=2pt, topsep=0pt, parsep=0pt]
    \item A joint scanpath-trajectory diffusion framework that couples both modalities via channel-level concatenation, enabling end-to-end training that exploits their complementarity without additional architectural complexity.
    \item A CRPS-based metrics family that generalizes any existing sequence similarity measure into a proper probabilistic scoring rule for evaluating the distributional accuracy and diversity of generative gaze models.
    \item State-of-the-art performance on free-viewing and task-driven benchmarks, with ablations confirming the contribution of joint training and the informativeness of the proposed metrics in distinguishing models that capture human gaze variability from those that do not.
\end{itemize}

\section{Related Works}
\label{sec:related_works}

\paragraph{Scanpath Generation.} Saliency prediction has a long history in visual perception research~\cite{Treisman1980, koch1987shifts, findlay2001visual, yarbus2013eye, zelinsky2008theory}, and following Itti's computational model~\cite{itti1998saliency} became an active area of study in computer vision~\cite{berg2009free, borji2012state, li2014secrets, jiang2015salicon, jetley2016end, kruthiventi2017deepfix, cornia2018predicting, kummerer2014deepgazeI, linardos2021deepgazeIIE, Lou2022transalnet, Hosseini2025sum}. Saliency maps, however, capture only where attention is likely to land, not the temporal order in which a viewer visits those locations, motivating a shift to scanpath generation: given an image and an optional target category for task-driven search, the model predicts an ordered sequence of fixations. Early approaches sampled fixations from hand-crafted~\cite{itti1998saliency,koch1987shifts} or deep~\cite{kummerer2014deepgazeI,linardos2021deepgazeIIE,cornia2018predicting} saliency maps, while later work modeled the fixation sequence directly. Recurrent and reinforcement-learning models~\cite{assens2018pathgan,sun2019iorroilstm,chen2021predictingcocofreeview} predict scanpaths autoregressively, and DeepGaze~III~\cite{kummerer2022deepgaze} couples a saliency backbone with a fixation density model. Transformer-based models~\cite{mondal2023gazeformer,yang2024hat,jiang2024eyeformer} extend this to longer-range dependencies and unified free-viewing/visual-search formulations, while individualized variants~\cite{chen2024beyond,chen2024gazexplain,damico2024tppgaze} condition on observer or task identity. Most of these models produce a single scanpath per stimulus and approximate average viewer behavior, leaving inter-subject variability unmodeled. Diffusion-based scanpath generation has only recently emerged: ScanDL~\cite{bolliger2023scandl} models reading scanpaths over text, and ScanDiff~\cite{scandiff} extends this to natural images with textual conditioning for task-driven search. These models capture stochasticity at the fixation level but still supervise on scanpaths and discard the underlying continuous gaze signal.

\paragraph{Trajectory Generation.} A second line of work supervises generative models on the raw eye-tracking trajectory, retaining the timing and geometry of saccades and the high-frequency dynamics during fixations. DiffEye~\cite{kara2025diffeye} is, to our knowledge, the first diffusion-based model trained directly on continuous trajectories from natural images, and reports gains on scanpath-level metrics after converting the generated trajectory into a fixation sequence. Related efforts include DiffEyeSyn~\cite{jiao2024diffeyesyn}, which synthesizes user-specific high-frequency eye movements, and an autoregressive diffusion model for gaze on 360$^{\circ}$ panoramic content~\cite{jiao2026diffgaze}. These methods establish that continuous gaze can be generated, but treat the trajectory as the sole training signal and do not exploit the discrete fixation structure that scanpath-based models make explicit. Our work bridges these two lines by training a single diffusion model jointly on trajectories and scanpaths, so that fine-grained dynamics and fixation-level structure inform a shared latent representation.
\section{Continuous Gaze Generation via Joint Trajectory-Scanpath Modeling}
\label{sec:method}

\subsection{Problem Formulation}

Human visual attention can be represented as a stochastic spatio-temporal process conditioned on a visual stimulus and, optionally, a viewing task. To capture this process, we define a dataset $\mathcal{D}$ comprising $N$ visual stimuli. We use the subscript $i \in \{1, \dots, N\}$ to index the specific image $I_i \in \mathbb{R}^{H \times W \times 3}$, and the superscript $j \in \{1, \dots, J_i\}$ to index the $J_i$ distinct human subjects who viewed it. The complete dataset is
\begin{equation}
    \mathcal{D} = \left\{ \left( I_i, c_i, \{R_i^j, S_i^j\}_{j=1}^{J_i} \right) \right\}_{i=1}^{N}, \quad 
    R_i^j = \{ (x_t^{\mathrm{tr}}, y_t^{\mathrm{tr}}) \}_{t=1}^{T_i^j}, \quad
    S_i^j = \{ (x_m^{\mathrm{sp}}, y_m^{\mathrm{sp}}, \Delta t_m^{\mathrm{sp}}) \}_{m=1}^{M_i^j}
\end{equation}
where $c_i$ is the optional task-related conditioning variable (such as a target category in visual search), $T_i^j$ is the number of recorded time steps in the raw trajectory $R_i^j$, and $M_i^j$ is the number of fixations in the scanpath $S_i^j$, with $(x_m^{\mathrm{sp}}, y_m^{\mathrm{sp}})$ the spatial centroid and $\Delta t_m^{\mathrm{sp}}$ the duration of the $m$-th fixation. The superscripts $\mathrm{tr}$ and $\mathrm{sp}$ distinguish trajectory and scanpath coordinates throughout.

Our primary goal is to learn a generative model of human gaze behavior. To train such models, previous approaches typically rely on either continuous trajectories $R$ or discrete scanpaths $S$ in isolation. In contrast, our framework utilizes both $R$ and $S$ jointly during training to capture their complementary information. Ultimately, we obtain a learned generative distribution $p_\theta(R \mid I, c)$ that, conditioned on a new visual stimulus $I$ and an optional task variable $c$, produces a continuous trajectory $R$, which is then postprocessed to yield the corresponding discrete scanpath $S$.

\subsection{\method}
\begin{figure*}[t!]
    \centering
    \includegraphics[width=\linewidth]{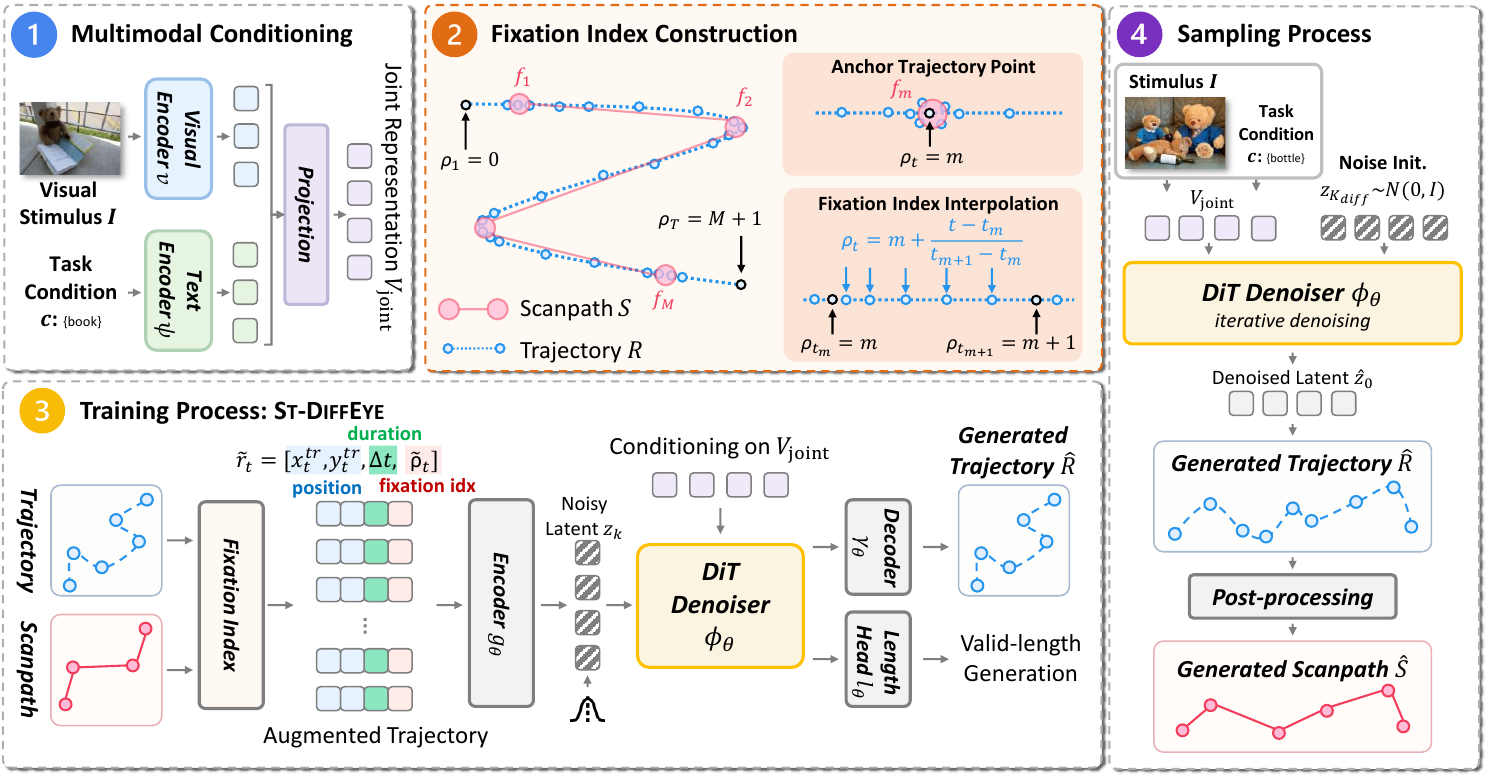}
    \caption{\textbf{Overview of \method.} (1) The visual stimulus $I$ and task condition $c$ are encoded by $v$ and $\psi$ and fused into a joint representation $V_{\mathrm{joint}}$. (2) Each trajectory point is assigned a continuous fixation index $\tilde{\rho}_t$ by anchoring fixations to their nearest trajectory points and linearly interpolating between anchors. (3) The augmented trajectory tokens are denoised by a DiT $\phi_\theta$ conditioned on $V_{\mathrm{joint}}$, then decoded into a trajectory with a predicted valid length. (4) At sampling time, the generated trajectory $\hat{R}$ is post-processed via fixation extraction to produce the final scanpath $\hat{S}$.}
    \label{fig:method}
\end{figure*}

\paragraph{Overview.}
Motivated by the capacity of DiTs~\cite{peebles2023scalabledit}, and unlike the U-Net backbone of prior gaze models~\cite{kara2025diffeye}, we follow ScanDiff~\cite{scandiff} in adopting a DiT and adapt it to operate directly on continuous trajectory data. To guide the generative process, we extract contextual features from the visual stimulus via a visual backbone $v(\cdot)$ and task-specific conditions through a text encoder $\psi(\cdot)$, fusing them into a joint multimodal representation $V_{\text{joint}}\in\mathbb{R}^{hw\times d}$. Rather than treating continuous trajectories and discrete scanpaths in isolation, we bridge them by augmenting each trajectory token with an additional channel that encodes the higher-level scanpath structure through a continuous \emph{fixation index}. The augmented sequence is projected into a latent token space $z_0=g_{\theta}(\tilde R)\in\mathbb{R}^{T\times d}$ and corrupted through a forward diffusion process. The denoising network $\phi_{\theta}$ models spatial and temporal dependencies of the gaze sequence and fuses the noisy gaze tokens $z_k$ with the multimodal condition $V_{\text{joint}}$ using self/cross-attention. Finally, to account for the natural variation in human gaze duration, the predicted clean latent $\hat{z}_0=\phi_{\theta}(z_k,V_{\text{joint}})$ is processed by a decoder $\gamma_{\theta}$ alongside a length-prediction head $l_{\theta}$ that estimates token validity, enabling variable-length generation. An overview is shown in \Cref{fig:method}.

\paragraph{Fixation Index.}
While continuous trajectories provide rich spatio-temporal dynamics, scanpaths encode higher-level semantic structure through discrete fixations. We exploit the natural spatial alignment between these two modalities to inject scanpath information into our trajectory modeling pipeline. Intuitively, trajectories can be viewed as a high-resolution signal, whereas scanpaths represent a compressed, low-resolution abstraction that preserves the most salient information. In particular, fixations correspond to semantically meaningful regions, while the intermediate saccades are more stochastic and less structured. Our goal is therefore to regularize these intermediate trajectory segments by aligning them with fixation-level structure.

To this end, we introduce a simple yet effective representation, termed the \emph{fixation index}, which assigns each trajectory point a continuous value indicating its relative position within the scanpath. Dropping the per-image and per-subject indices $(i, j)$ for clarity, we work with a single trajectory $R = \{(x_t^{\mathrm{tr}}, y_t^{\mathrm{tr}})\}_{t=1}^T$ and its corresponding scanpath $S = \{(x_m^{\mathrm{sp}}, y_m^{\mathrm{sp}}, \Delta t_m^{\mathrm{sp}})\}_{m=1}^M$ as defined above; the durations $\Delta t_m^{\mathrm{sp}}$ are not used in this construction. We first identify a set of \emph{anchor trajectory points} in the trajectory by aligning each fixation to its nearest trajectory point:
\begin{equation}
    t_m = \arg\min_{t} \big\|(x_t^{\mathrm{tr}}, y_t^{\mathrm{tr}}) - (x_m^{\mathrm{sp}}, y_m^{\mathrm{sp}})\big\|_2, \quad m=1,\dots,M,
\end{equation}
while enforcing temporal consistency $1 \leq t_1 < t_2 < \dots < t_M \leq T$. We additionally define boundary anchors $t_0=1$ and $t_{M+1}=T$. We then construct a fixation index sequence $\rho=\{\rho_t\}_{t=1}^{T}$, where each anchor point is assigned its corresponding index:
\begin{equation}
    \rho_{t_m}=m,\quad m=1,\dots,M, \qquad \rho_{t_0}=0,\quad \rho_{t_{M+1}}=M+1.
\end{equation}
For trajectory points between two consecutive anchors $t_m$ and $t_{m+1}$, we assign values by linear \emph{fixation index interpolation}:
\begin{equation}
    \rho_t = m + \frac{t-t_m}{t_{m+1}-t_m}, \quad t\in[t_m,t_{m+1}], \quad m=0,\dots,M.
\end{equation}
This yields a continuous-valued index that encodes the progression of each trajectory point within the fixation sequence. Values close to $m$ indicate proximity to the $m$-th fixation, while intermediate values such as $m+0.5$ correspond to mid-saccade transitions. Finally, we normalize the index as $\tilde{\rho}_t=\rho_t/(M+1)$ and include the per-sample time step $\Delta t = 1/\text{sample rate}$ to encode the temporal resolution of the trajectory. The final augmented trajectory token is
\begin{equation}
    \tilde{r}_t = [x_t^{\mathrm{tr}},\, y_t^{\mathrm{tr}},\, \Delta t,\, \tilde{\rho}_t].
\end{equation}
This augmentation provides each trajectory point with explicit global context about its role in the overall gaze process, enabling the model to better structure stochastic saccadic motion while preserving the semantic organization induced by fixations.

\paragraph{Training.}
Given the augmented trajectory sequence $\tilde{R}=\{\tilde{r}_t\}_{t=1}^{T}$ with $\tilde{r}_t=[x_t^{\mathrm{tr}}, y_t^{\mathrm{tr}}, \Delta t, \tilde{\rho}_t]$, we first project the sequence into a latent representation $z_0=g_{\theta}(\tilde{R})$. Following the Denoising Diffusion Probabilistic Model (DDPM) framework~\cite{ho2020denoisingddpm}, we define a forward diffusion process that gradually adds Gaussian noise to $z_0$ over $K$ steps according to a predefined variance schedule $\beta_1, \dots, \beta_K$, given by $q(z_k \mid z_{k-1}) = \mathcal{N}(z_k;\, \sqrt{1 - \beta_k}\, z_{k-1},\, \beta_k I)$. This formulation allows us to sample the noisy latent $z_k$ at an arbitrary step $k$ directly from $z_0$ in closed form as $q(z_k \mid z_0) = \mathcal{N}(z_k;\, \sqrt{\bar{\alpha}_k}\, z_0,\, (1 - \bar{\alpha}_k) I)$, where $\alpha_k = 1 - \beta_k$ and $\bar{\alpha}_k = \prod_{i=1}^k \alpha_i$. The denoising network $\phi_{\theta}$ is trained to reverse this process and recover $z_0$, conditioned on the diffusion step $k$ and the multimodal stimulus representation $V_{\mathrm{joint}}$. Following ScanDiff~\cite{scandiff}, we adopt the same training objective (see Appendix for implementation details), which minimizes a weighted sum of four components: 
\begin{equation}
\mathcal{L} = \mathcal{L}_{\mathrm{diff}} + \lambda_{\mathrm{recon}}\mathcal{L}_{\mathrm{recon}} + \lambda_{\mathrm{val}}\mathcal{L}_{\mathrm{val}} + \lambda_{\mathrm{reg}}\mathcal{L}_{\mathrm{reg}}.
\end{equation}
The diffusion loss $\mathcal{L}_{\mathrm{diff}}$ supervises the latent denoising process. The reconstruction loss $\mathcal{L}_{\mathrm{recon}}$ ensures that the decoded output $\hat{R}=\gamma_{\theta}(\hat{z}_0)$ accurately maps the predicted clean latent back to the augmented trajectory space. The validity loss $\mathcal{L}_{\mathrm{val}}$ trains the length-prediction head to distinguish valid trajectory tokens from padding elements. The terminal regularization loss $\mathcal{L}_{\mathrm{reg}}$ encourages the distribution of the final diffusion state $z_K$ to match an isotropic Gaussian prior.

\paragraph{Sampling.}
Given a visual stimulus and task condition, we first sample $z_K\sim\mathcal{N}(0,I)$ and iteratively apply the learned reverse diffusion process conditioned on $V_{\mathrm{joint}}$ to obtain the predicted clean latent $\hat{z}_0$. The decoder $\gamma_{\theta}$ then maps $\hat{z}_0$ back to the augmented trajectory representation $\hat{R}=\{\hat{r}_t\}_{t=1}^{T}$, where each predicted token is $\hat{r}_t=[\hat{x}_t^{\mathrm{tr}}, \hat{y}_t^{\mathrm{tr}}, \Delta t, \hat{\rho}_t]$. Since the fixation index is introduced only as an auxiliary structural signal for trajectory generation, we discard the predicted index channel $\hat{\rho}_t$ after decoding and retain the continuous trajectory $\{(\hat{x}_t^{\mathrm{tr}}, \hat{y}_t^{\mathrm{tr}})\}_{t=1}^{T}$ together with the fixed step size $\Delta t$. Finally, we convert the generated trajectory into a discrete scanpath using the fixation extraction algorithm provided by~\cite{Judd_2009mit1003}. By sampling different initial noise sequences $z_K$ for the same stimulus, the model produces multiple plausible trajectories and scanpaths that reflect the stochastic variability of human gaze. More implementation details are shown in Appendix \Cref{sec:scandiff_preliminaries,sec:implementation_details}.

\begin{figure*}[t]
    \centering
    \includegraphics[width=1\textwidth]{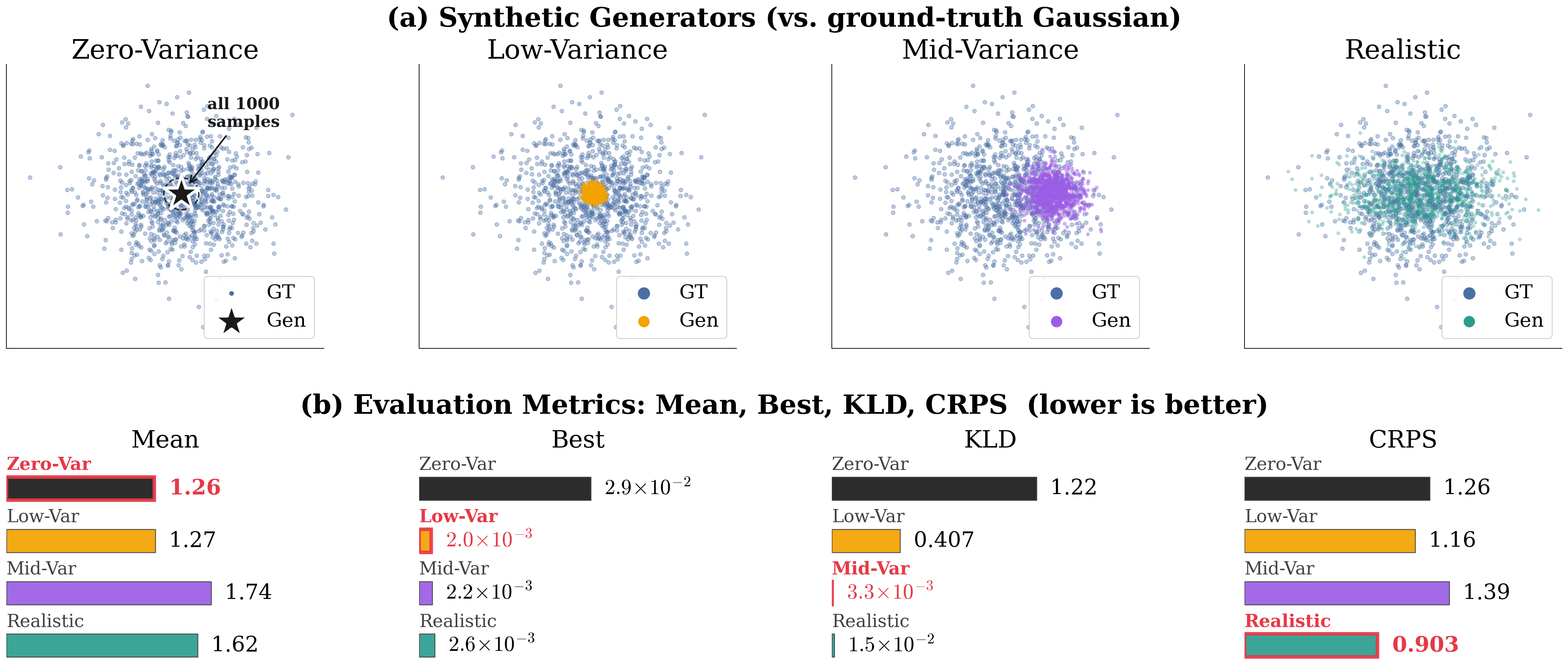} 
    \caption{\textbf{Failure modes of mean, best, and KLD evaluation protocols.} (a) Four generators evaluated against the same ground-truth Gaussian (blue): zero-variance, low-variance, mid-variance, and realistic. (b) Scores under each protocol (lower is better). Each non-CRPS protocol is won by a different degenerate generator (mean: zero-variance; best: low-variance; KLD: mid-variance), while only CRPS correctly identifies the realistic generator.}
    \label{fig:crps}
\end{figure*}
\subsection{CRPS-based Evaluation}
\label{sec:crps}

Generative gaze models are commonly evaluated under two point-estimate protocols~\cite{kara2025diffeye, jiao2026diffgaze}: a \emph{mean} protocol averaging a similarity metric over all generated--GT pairs, and a \emph{best} protocol reporting only the closest generated sample per GT. Both have well-known failure modes, illustrated on synthetic 2D Gaussian generators in \Cref{fig:crps}: the mean protocol rewards mode collapse (zero-variance generator), while the best protocol rewards low-variance generators that produce at least one close sample per GT but otherwise ignore the multi-modality of human gaze (low-variance generator). ScanDiff~\cite{scandiff} and TPP-Gaze~\cite{damico2024tppgaze} instead adopt a Kullback-Leibler divergence (KLD) protocol over score histograms, but the same histogram can arise from very different sample distributions (mid-variance generator).

We therefore propose a distribution-aware protocol based on the Continuous Ranked Probability Score (CRPS)~\cite{Gneiting01032007crps}, inspired by~\cite{mammadov2026variationalcrps}, which evaluates the predictive distribution directly by jointly measuring accuracy and diversity. Given ground-truth samples $\mathbf{x}=\{x^{(i)}\}_{i=1}^N$ and generated samples $\hat{\mathbf{x}}=\{\hat{x}^{(j)}\}_{j=1}^J$,
\begin{equation}
    \mathrm{CRPS}(\hat{\mathbf{x}},\mathbf{x},d) 
    = \frac{1}{N}\sum_{i=1}^N\left[
        \frac{1}{J}\sum_{j=1}^J d\!\left(\hat{x}^{(j)},x^{(i)}\right) 
        - \frac{1}{2J^2}\sum_{j=1}^J\sum_{k=1}^J d\!\left(\hat{x}^{(j)},\hat{x}^{(k)}\right)
    \right].
\end{equation}
The first term pulls generated samples toward the ground truth, while the second penalizes collapse through their pairwise distances; the score is minimized only when the generated distribution is both close to the ground truth and as spread out as it is. Since $d(\cdot,\cdot)$ can be any scanpath distance or similarity (L2 distance in our illustration), CRPS provides a unified evaluation that resolves the failure modes of mean, best, and KLD protocols, with only the realistic generator winning.

\section{Experiments}
\label{sec:experiments}

\subsection{Evaluation Setup}

\begin{figure*}[t]
    \centering
    \includegraphics[width=\linewidth]{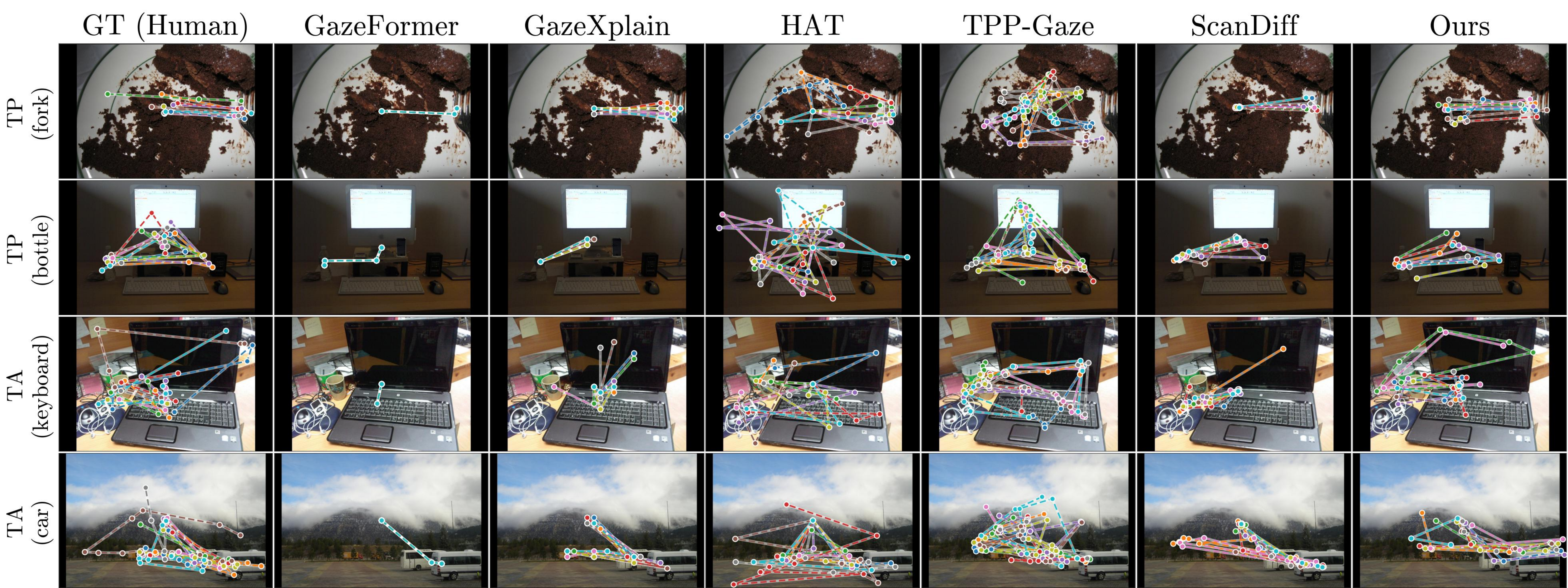} \\
    \includegraphics[width=\linewidth]{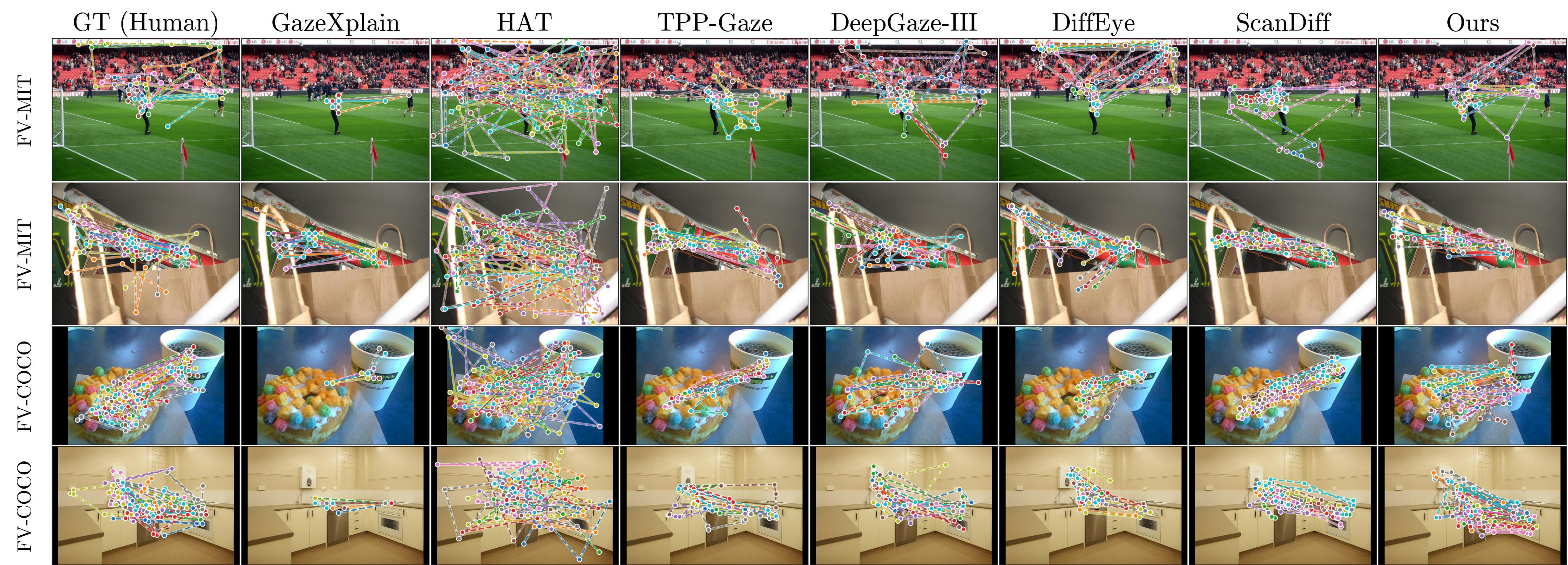}
    \caption{\textbf{Qualitative comparison with baselines.} Generated scanpaths on COCO-Search18 (top, target-present and target-absent rows) and the free-viewing benchmarks COCO-FreeView and MIT1003 (bottom).}
    \label{fig:compare_main}
\end{figure*}

\paragraph{Datasets and Preprocessing.}
We evaluate our framework on two canonical gaze generation tasks: task-driven visual search and free-viewing. Each sample comprises an image stimulus, a discrete scanpath, and its corresponding continuous gaze trajectory, with visual search samples additionally including a target object description. 
For visual search, we use \textbf{COCO-Search18}~\cite{chen2021cocosearch18} (6,202 images, 18 target categories, 10 subjects, $\sim$300K fixations). For free-viewing, we use \textbf{COCO-FreeView}~\cite{chen2021predictingcocofreeview} (the same COCO images under a free-viewing protocol, >800K fixations) and \textbf{MIT1003}~\cite{Judd_2009mit1003} (1,003 images, 15 subjects).
Because only discrete scanpaths are publicly available for the COCO datasets, we processed the original EDF files to extract the raw continuous trajectories. To align these newly extracted trajectories with the public scanpaths, we trim the leading and trailing trajectory segments that fall outside a 20-pixel radius of the initial and final fixations. All trajectories are then downsampled to 240 Hz and padded or truncated to a maximum of 1500 points, while scanpaths are bounded to a maximum of 16 fixations. Image preprocessing, text feature extraction, and train/test splits strictly follow ScanDiff~\cite{scandiff}.

\paragraph{Metrics.}
We primarily evaluate models using the CRPS protocol from \Cref{sec:crps}, instantiated with two complementary metric families. Following DiffEye~\cite{kara2025diffeye}, we use \emph{distance-based} metrics that quantify geometric and temporal misalignment: Levenshtein Distance (LD), Discrete Fr\'echet Distance (DFD), Dynamic Time Warping (DTW), and Time Delay Embedding (TDE). Following ScanDiff~\cite{scandiff}, we use \emph{similarity-based} metrics: MultiMatch (MM), ScanMatch (SM), Sequence Score (SS), and Semantic Sequence Score (SSS). SSS is reported only for visual search since it relies on target-category supervision. For SM, SS, and SSS, we additionally report duration-aware variants ($\dagger$) that incorporate fixation durations. KLD, Mean, Best-protocol results are included in the appendix.

\paragraph{Baselines.}
We compare against representative gaze models grouped by modeling strategy. The first group covers non-diffusion scanpath predictors: GazeFormer~\cite{mondal2023gazeformer}, a task-conditioned transformer producing a single sequence per stimulus; HAT~\cite{yang2024hat}, a transformer unifying top-down and bottom-up scanpath prediction; and DeepGaze~III~\cite{kummerer2022deepgaze}, a saliency-based probabilistic model for free-viewing. The second adds extra conditioning or stochasticity: GazeXplain~\cite{chen2024gazexplain} uses language-based supervision, and TPP-Gaze~\cite{damico2024tppgaze} models fixation arrivals with neural temporal point processes. The third comprises diffusion-based generative models: ScanDiff~\cite{scandiff} generates scanpaths with a task-conditioned diffusion transformer, and DiffEye~\cite{kara2025diffeye} generates continuous trajectories and converts them into scanpaths for free-viewing. ScanDiff and DiffEye are most directly comparable to our setting, capturing stochasticity at the fixation and trajectory levels respectively, while we model both within a single diffusion process.

\subsection{Analysis and Discussion}
\begin{table*}[t]
\centering
\caption{\textbf{CRPS metrics on free-viewing and visual-search datasets.} Datasets: COCO-FV (COCO-FreeView), MIT-FV (MIT1003), COCO-TP (COCO-Search18, Target Present), COCO-TA (COCO-Search18, Target Absent). ScanMatch (SM), Sequence Score (SS), and Semantic Sequence Score (SSS); $\dagger$ marks the duration-aware variant. \textbf{Bold}: best; \underline{underlined}: second best within each dataset and comparison block. Dashes indicate methods that do not predict durations, or metrics that do not apply (SSS is undefined for free-viewing).}
\label{tab:crps_all}
\vspace{0.5em}
\setlength{\tabcolsep}{3pt}
\renewcommand{\arraystretch}{1.05}
\resizebox{\textwidth}{!}{%
\begin{tabular}{l l l ccccc cccc cc}
\toprule
\textbf{Dataset} & \textbf{Comparison} & \textbf{Method}
& LD$\downarrow$ & DFD$\downarrow$ & DTW$\downarrow$ & TDE$\downarrow$ & MM$\uparrow$
& SM$\uparrow$ & SM$^{\dagger}\uparrow$ & SS$\uparrow$ & SS$^{\dagger}\uparrow$ & SSS$\uparrow$ & SSS$^{\dagger}\uparrow$ \\
\midrule
\multirow{9}{*}{\rotatebox[origin=c]{90}{\textsc{COCO-FV}}}
& \multirow{7}{*}{\textit{Scanpath}}
& GazeXplain~\cite{chen2024gazexplain} & 95.512 & 168.087 & 1418.642 & 0.059 & 0.323 & $-$0.232 & $-$0.231 & 0.099 & $-$0.220 & -- & -- \\
& & HAT~\cite{yang2024hat} & 50.825 & 106.551 & 944.485 & 0.052 & \underline{0.413} & \underline{0.099} & -- & $-$0.170 & -- & -- & -- \\
& & TPP-Gaze~\cite{damico2024tppgaze} & 64.219 & 109.774 & 1076.181 & \textbf{0.035} & 0.385 & 0.055 & 0.022 & 0.163 & 0.019 & -- & -- \\
& & DeepGaze III~\cite{kummerer2022deepgaze} & 66.587 & 104.779 & 998.289 & \underline{0.040} & \textbf{0.414} & 0.033 & -- & $-$0.235 & -- & -- & -- \\
& & DiffEye~\cite{kara2025diffeye} & 56.368 & 121.355 & 1116.594 & 0.051 & 0.370 & 0.025 & $-$0.018 & 0.197 & $-$0.029 & -- & -- \\
& & ScanDiff~\cite{scandiff} & \textbf{41.856} & \underline{104.744} & \underline{910.433} & 0.049 & 0.387 & \textbf{0.102} & \underline{0.089} & \textbf{0.274} & \underline{0.075} & -- & -- \\
& & \cellcolor{gray!10}\method & \cellcolor{gray!10}\underline{47.108} & \cellcolor{gray!10}\textbf{103.009} & \cellcolor{gray!10}\textbf{851.792} & \cellcolor{gray!10}0.048 & \cellcolor{gray!10}0.377 & \cellcolor{gray!10}0.089 & \cellcolor{gray!10}\textbf{0.108} & \cellcolor{gray!10}\underline{0.206} & \cellcolor{gray!10}\textbf{0.088} & \cellcolor{gray!10}-- & \cellcolor{gray!10}-- \\
\cmidrule(l){2-14}
& \multirow{2}{*}{\textit{Trajectory}}
& DiffEye~\cite{kara2025diffeye} & \underline{56.349} & \underline{121.493} & \underline{1120.990} & \underline{0.051} & \underline{0.289} & \underline{0.025} & $-$0.196 & \textbf{0.187} & $-$0.147 & -- & -- \\
& & \cellcolor{gray!10}\method & \cellcolor{gray!10}\textbf{53.487} & \cellcolor{gray!10}\textbf{105.121} & \cellcolor{gray!10}\textbf{957.552} & \cellcolor{gray!10}\textbf{0.044} & \cellcolor{gray!10}\textbf{0.299} & \cellcolor{gray!10}\textbf{0.028} & \cellcolor{gray!10}\textbf{$-$0.097} & \cellcolor{gray!10}\underline{0.102} & \cellcolor{gray!10}\textbf{$-$0.090} & \cellcolor{gray!10}-- & \cellcolor{gray!10}-- \\
\midrule
\multirow{9}{*}{\rotatebox[origin=c]{90}{\textsc{MIT-FV}}}
& \multirow{7}{*}{\textit{Scanpath}}
& GazeXplain~\cite{chen2024gazexplain} & 32.979 & 115.568 & 640.062 & 0.071 & 0.350 & 0.009 & 0.049 & 0.285 & 0.036 & -- & -- \\
& & HAT~\cite{yang2024hat} & 34.782 & 119.965 & 667.106 & 0.084 & \underline{0.399} & 0.080 & -- & $-$0.153 & -- & -- & -- \\
& & TPP-Gaze~\cite{damico2024tppgaze} & 26.109 & 103.796 & 502.235 & \textbf{0.052} & 0.378 & 0.112 & 0.094 & 0.255 & 0.089 & -- & -- \\
& & DeepGaze III~\cite{kummerer2022deepgaze} & 26.787 & 102.286 & \underline{465.898} & \underline{0.058} & \textbf{0.419} & \textbf{0.130} & -- & $-$0.107 & -- & -- & -- \\
& & DiffEye~\cite{kara2025diffeye} & 27.530 & 111.506 & 516.821 & 0.062 & 0.377 & 0.112 & \underline{0.125} & \underline{0.328} & 0.121 & -- & -- \\
& & ScanDiff~\cite{scandiff} & \underline{25.189} & \underline{102.216} & 496.327 & \underline{0.058} & 0.379 & 0.111 & 0.118 & \textbf{0.342} & \underline{0.128} & -- & -- \\
& & \cellcolor{gray!10}\method & \cellcolor{gray!10}\textbf{24.961} & \cellcolor{gray!10}\textbf{100.903} & \cellcolor{gray!10}\textbf{456.985} & \cellcolor{gray!10}\underline{0.058} & \cellcolor{gray!10}0.380 & \cellcolor{gray!10}\underline{0.129} & \cellcolor{gray!10}\textbf{0.145} & \cellcolor{gray!10}0.321 & \cellcolor{gray!10}\textbf{0.146} & \cellcolor{gray!10}-- & \cellcolor{gray!10}-- \\
\cmidrule(l){2-14}
& \multirow{2}{*}{\textit{Trajectory}}
& DiffEye~\cite{kara2025diffeye} & \textbf{26.351} & \underline{111.252} & \underline{496.530} & \underline{0.061} & \textbf{0.378} & \textbf{0.123} & \underline{0.129} & \textbf{0.333} & \underline{0.124} & -- & -- \\
& & \cellcolor{gray!10}\method & \cellcolor{gray!10}\underline{26.589} & \cellcolor{gray!10}\textbf{108.085} & \cellcolor{gray!10}\textbf{486.137} & \cellcolor{gray!10}\textbf{0.060} & \cellcolor{gray!10}\underline{0.373} & \cellcolor{gray!10}\underline{0.114} & \cellcolor{gray!10}\textbf{0.139} & \cellcolor{gray!10}\underline{0.287} & \cellcolor{gray!10}\textbf{0.133} & \cellcolor{gray!10}-- & \cellcolor{gray!10}-- \\
\midrule
\multirow{6}{*}{\rotatebox[origin=c]{90}{\textsc{COCO-TP}}}
& \multirow{6}{*}{\textit{Scanpath}}
& GazeFormer~\cite{mondal2023gazeformer} & 15.115 & 181.157 & 284.039 & 0.099 & 0.327 & 0.044 & 0.008 & \textbf{0.540} & 0.052 & 0.001 & $-$0.045 \\
& & GazeXplain~\cite{chen2024gazexplain} & 13.808 & 184.248 & 283.833 & 0.094 & 0.334 & 0.066 & 0.066 & \underline{0.535} & 0.102 & 0.039 & 0.037 \\
& & HAT~\cite{yang2024hat} & 10.784 & \underline{122.199} & 227.450 & 0.080 & \textbf{0.398} & \underline{0.201} & -- & 0.197 & -- & \underline{0.185} & -- \\
& & TPP-Gaze~\cite{damico2024tppgaze} & 25.299 & 179.339 & 476.733 & 0.121 & 0.348 & $-$0.014 & $-$0.049 & 0.180 & $-$0.088 & $-$0.036 & $-$0.095 \\
& & ScanDiff~\cite{scandiff} & \underline{9.653} & 124.455 & \underline{219.855} & \underline{0.076} & 0.371 & 0.165 & \underline{0.104} & 0.521 & \underline{0.148} & 0.164 & \underline{0.089} \\
& & \cellcolor{gray!10}\method & \cellcolor{gray!10}\textbf{8.716} & \cellcolor{gray!10}\textbf{110.038} & \cellcolor{gray!10}\textbf{196.631} & \cellcolor{gray!10}\textbf{0.074} & \cellcolor{gray!10}\underline{0.381} & \cellcolor{gray!10}\textbf{0.206} & \cellcolor{gray!10}\textbf{0.146} & \cellcolor{gray!10}0.504 & \cellcolor{gray!10}\textbf{0.180} & \cellcolor{gray!10}\textbf{0.211} & \cellcolor{gray!10}\textbf{0.134} \\
\midrule
\multirow{6}{*}{\rotatebox[origin=c]{90}{\textsc{COCO-TA}}}
& \multirow{6}{*}{\textit{Scanpath}}
& GazeFormer~\cite{mondal2023gazeformer} & 32.439 & 188.240 & 622.325 & 0.100 & 0.308 & $-$0.161 & $-$0.171 & 0.293 & $-$0.170 & $-$0.141 & $-$0.160 \\
& & GazeXplain~\cite{chen2024gazexplain} & 28.302 & 152.507 & 514.880 & 0.074 & 0.350 & $-$0.009 & 0.001 & \textbf{0.347} & 0.010 & 0.002 & 0.009 \\
& & HAT~\cite{yang2024hat} & 24.371 & \underline{117.266} & 439.956 & \textbf{0.064} & \textbf{0.399} & \underline{0.095} & -- & $-$0.051 & -- & 0.100 & -- \\
& & TPP-Gaze~\cite{damico2024tppgaze} & 27.822 & 125.431 & 545.535 & 0.085 & 0.370 & 0.035 & 0.007 & 0.186 & $-$0.001 & 0.038 & $-$0.006 \\
& & ScanDiff~\cite{scandiff} & \underline{21.252} & 119.689 & \underline{426.350} & \underline{0.069} & 0.378 & 0.086 & \underline{0.077} & \underline{0.305} & \underline{0.090} & \underline{0.109} & \underline{0.092} \\
& & \cellcolor{gray!10}\method & \cellcolor{gray!10}\textbf{21.128} & \cellcolor{gray!10}\textbf{115.683} & \cellcolor{gray!10}\textbf{410.444} & \cellcolor{gray!10}\textbf{0.064} & \cellcolor{gray!10}\underline{0.381} & \cellcolor{gray!10}\textbf{0.110} & \cellcolor{gray!10}\textbf{0.095} & \cellcolor{gray!10}0.288 & \cellcolor{gray!10}\textbf{0.109} & \cellcolor{gray!10}\textbf{0.130} & \cellcolor{gray!10}\textbf{0.109} \\
\bottomrule
\end{tabular}}
\end{table*}

\paragraph{Main results.}
We report CRPS-based results across all four datasets in \Cref{tab:crps_all}.
\textbf{Visual search.} On COCO-Search18, \method achieves the strongest overall performance in both target-present and target-absent regimes, obtaining the best score on 9 of 11 metrics in COCO-TP and 8 of 11 in COCO-TA. The largest improvements over ScanDiff, the closest diffusion-based scanpath baseline, are on the distance-based metrics (DFD, DTW, TDE) and the duration-aware variants (SM$^\dagger$, SS$^\dagger$, SSS$^\dagger$), suggesting that trajectory information provides complementary fine-grained temporal cues beyond discrete fixation sequences.
\textbf{Free-viewing.} On COCO-FreeView and MIT1003, \method again leads the scanpath comparison block, ranking first on 4 of 9 and 5 of 9 metrics respectively, with consistent gains on DFD, DTW, and the duration-aware similarity metrics. The improvements on duration-aware metrics on both free-viewing benchmarks are particularly notable: \method exceeds ScanDiff on every duration-aware variant, indicating that the joint trajectory channel captures temporal alignment that scanpath-only supervision discards. While DeepGaze~III and HAT remain competitive on MM and non-duration-aware SM, our consistent gains on trajectory-sensitive and duration-aware metrics show that joint modeling better captures the distributional properties of free-viewing gaze.
\textbf{Trajectory comparison.} Beyond scanpath-level evaluation, the trajectory comparison block shows that \method also improves over DiffEye, the only continuous-trajectory baseline, on COCO-FreeView (best on 7 of 9 metrics) and on MIT1003 (best on 5 of 9 metrics). This confirms that joint scanpath supervision benefits not only the extracted scanpaths but also the underlying continuous trajectories.
Overall, jointly modeling scanpaths and trajectories yields consistent gains across both task regimes and both evaluation granularities, with the strongest improvements on the trajectory-sensitive and duration-aware metrics that prior scanpath-only baselines are not designed to capture.
More results under the mean, best, and KLD evaluation protocols are provided in Appendix \Cref{sec:additional_protocols}.

\paragraph{Ablation Study.}

\begin{table*}[t]
\centering
\caption{\textbf{Ablation studies on COCO-Search18 (Target Present).} We isolate four design choices: (a) input modality, (b) fixation-index interpolation, (c) concatenation stage, and (d) fixation-index encoding. The shaded row is our full model, combining the default of each group (trajectory + scanpath input, uniform interpolation, raw input channel concatenation, raw scalar encoding). \textbf{Bold} marks the best in each column.}
\label{tab:ablation}
\vspace{0.5em}
\scriptsize
\setlength{\tabcolsep}{3pt}
\renewcommand{\arraystretch}{1.0}
\resizebox{0.95\linewidth}{!}{%
\begin{tabular}{@{} l l *{11}{c} @{}}
\toprule
\textbf{Group} & \textbf{Variant}
& LD$\downarrow$ & DFD$\downarrow$ & DTW$\downarrow$ & TDE$\downarrow$
& MM$\uparrow$ & SM$\uparrow$ & SM$^{\dagger}\uparrow$ & SS$\uparrow$ & SS$^{\dagger}\uparrow$ & SSS$\uparrow$ & SSS$^{\dagger}\uparrow$ \\
\midrule
\multirow{2}{1.5cm}{\textit{(a) Input modality}}
& Scanpath only   & 9.653  & 124.455 & 219.855 & 0.076 & 0.371 & 0.165  & 0.104  & \textbf{0.521} & 0.148  & 0.164  & 0.089 \\
& Trajectory only & 9.180  & 121.767 & 237.234 & 0.090 & 0.372 & 0.164  & 0.112  & 0.445 & 0.160  & 0.178  & 0.110 \\
\midrule
\textit{(b) Interpolation}
& Distance-based  & 9.063  & 120.182 & 219.703 & 0.085 & 0.371 & 0.176  & 0.113  & 0.483 & 0.160  & 0.179  & 0.098 \\
\midrule
\multirow{2}{1.5cm}{\textit{(c) Concat. stage}}
& Latent-space    & 8.984  & 120.450 & 219.644 & 0.086 & 0.376 & 0.168  & 0.114  & 0.481 & 0.156  & 0.177  & 0.110 \\
& Embedding-space & 16.949 & 323.933 & 838.854 & 0.272 & 0.075 & $-$0.302 & $-$0.281 & $-$0.204 & $-$0.135 & $-$0.288 & $-$0.112 \\
\midrule
\multirow{2}{1.5cm}{\textit{(d) Index encoding}}
& Learnable       & 8.941  & 115.029 & 214.956 & 0.083 & 0.372 & 0.182  & 0.125  & 0.495 & 0.169  & 0.186  & 0.108 \\
& Sinusoidal      & 9.282  & 119.815 & 222.527 & 0.083 & 0.372 & 0.199  & 0.105  & 0.519 & 0.139  & 0.207  & 0.093 \\
\midrule
\rowcolor[HTML]{E8F0FE}
\multicolumn{2}{l}{\method} & \textbf{8.716} & \textbf{110.038} & \textbf{196.631} & \textbf{0.074} & \textbf{0.381} & \textbf{0.206} & \textbf{0.146} & 0.504 & \textbf{0.180} & \textbf{0.211} & \textbf{0.134} \\
\bottomrule
\end{tabular}%
}
\end{table*}

We conduct ablation studies on the COCO-Search18 target-present split to validate our design choices, as shown in \Cref{tab:ablation}. 
\textbf{Input modality.} The full model outperforms scanpath-only and trajectory-only variants on most metrics, confirming that fixation-level structure and continuous gaze dynamics provide complementary information. 
\textbf{Fixation index construction.} Uniform interpolation performs better than the distance-based alternative, which assigns each trajectory point a fixation index using an inverse-distance-weighted average of nearby fixation indices, suggesting that preserving temporal progression between fixations is more effective than relying only on spatial proximity. 
\textbf{Concatenation stage.} Injecting the fixation index as a raw input channel gives the best results, while latent-space concatenation is weaker and embedding-space conditioning substantially degrades performance. 
\textbf{Encoding type.} Using the raw scalar index generally outperforms using the index to retrieve learnable and sinusoidal embeddings, indicating that the continuous fixation index already provides sufficient alignment information. 
Overall, these results show that our simple joint input design with raw-channel fixation index conditioning is effective and robust.

\paragraph{Qualitative Evaluation.}

We visualize generated scanpaths in \Cref{fig:compare_main}. On COCO-Search18, our model produces accurate and task-consistent scanpaths in the target-present setting: the generated fixations reliably move toward the queried object and land within the target region, while still following the overall spatial pattern of human scanpaths. In the target-absent setting and free-viewing tasks, where gaze behavior is less goal-directed and more exploratory, our samples generally traverse image regions consistent with the ground-truth scanpath distribution. In contrast, several baselines either produce deterministic or low-diversity outputs, such as GazeFormer, GazeXplain, and ScanDiff, or fail to match the visual structure of human gaze patterns, as observed for HAT, TPP-Gaze, and DiffEye. These qualitative results suggest that our joint trajectory-scanpath modeling better captures both task relevance and human-like variability. More visualizations and statistical analysis of generation results are shown in Appendix \Cref{sec:additional_qualitative,sec:scanpath_stats}.



\section{Conclusion}
\label{sec:conclusion}

We present \method, a joint trajectory-scanpath diffusion framework for stochastic human gaze modeling. By aligning scanpaths with continuous trajectories through fixation-index channels, our method combines fixation-level structure with fine-grained gaze dynamics using only minimal input-output modifications. We also introduce CRPS-based metrics for evaluating both accuracy and diversity. Experiments on free-viewing and visual search benchmarks show that joint modeling outperforms single-representation baselines and achieves state-of-the-art performance, highlighting the complementarity of scanpaths and trajectories for generative gaze modeling. Limitations, broader societal impacts, and the licenses of assets used in this work are discussed in Appendix \Cref{sec:limitations,sec:broader_impacts,sec:licenses}.

\clearpage\newpage
{
    \small
    \bibliographystyle{unsrt}
    \bibliography{main}
}


\clearpage\newpage
\appendix

\begin{center}
    {\Large\bfseries Appendix}
\end{center}
\vspace{1em}
\startcontents[appendix]
\printcontents[appendix]{}{1}{\section*{Contents}\setcounter{tocdepth}{2}}
\vspace{1.5em}

\clearpage
\section{Additional Evaluation Protocols}
\label{sec:additional_protocols}
We report additional results under the mean, best-of-$N$, and KLD evaluation protocols.

\subsection{Mean Protocol}
\begin{table*}[ht!]
\centering
\caption{\textbf{Mean metrics across free-viewing and visual-search datasets.} Datasets: COCO-FV (COCO-FreeView), MIT-FV (MIT1003), COCO-TP (COCO-Search18, Target Present), COCO-TA (COCO-Search18, Target Absent). ScanMatch (SM), Sequence Score (SS), and Semantic Sequence Score (SSS); $\dagger$ marks the duration-aware variant. \textbf{Bold}: best; \underline{underlined}: second best within each dataset. Dashes indicate methods that do not predict durations, or metrics that do not apply (SSS is undefined for free-viewing).}
\label{tab:all_mean}
\setlength{\tabcolsep}{3pt}
\renewcommand{\arraystretch}{1.05}
\resizebox{\textwidth}{!}{%
\begin{tabular}{l l ccccc cccc cc}
\toprule
\textbf{Dataset} & \textbf{Method}
& LD$\downarrow$ & DFD$\downarrow$ & DTW$\downarrow$ & TDE$\downarrow$ & MM$\uparrow$
& SM$\uparrow$ & SM$^{\dagger}\uparrow$ & SS$\uparrow$ & SS$^{\dagger}\uparrow$ & SSS$\uparrow$ & SSS$^{\dagger}\uparrow$ \\
\midrule
\multirow{7}{*}{\rotatebox[origin=c]{90}{\textsc{COCO-FV}}}
& GazeXplain~\cite{chen2024gazexplain} & 98.724 & 208.090 & 1476.637 & 0.081 & 0.785 & 0.163 & 0.141 & 0.155 & 0.134 & -- & -- \\
& HAT~\cite{yang2024hat} & 76.518 & 204.100 & 1636.314 & 0.101 & 0.671 & 0.304 & -- & 0.289 & -- & -- & -- \\
& TPP-Gaze~\cite{damico2024tppgaze} & 84.113 & 185.634 & 1454.071 & 0.080 & 0.792 & 0.243 & 0.218 & 0.213 & 0.186 & -- & -- \\
& DeepGaze III~\cite{kummerer2022deepgaze} & 81.550 & 188.692 & \textbf{1359.499} & 0.085 & 0.677 & 0.279 & -- & 0.266 & -- & -- & -- \\
& DiffEye~\cite{kara2025diffeye} & 78.838 & \underline{181.144} & \underline{1406.843} & \underline{0.078} & \underline{0.803} & \underline{0.309} & 0.283 & 0.285 & 0.253 & -- & -- \\
& ScanDiff~\cite{scandiff} & \underline{78.396} & \textbf{171.134} & 1480.892 & \textbf{0.076} & \textbf{0.831} & \textbf{0.356} & \textbf{0.346} & \textbf{0.324} & \textbf{0.299} & -- & -- \\
\rowcolor{gray!10}
\cellcolor{white} & \method & \textbf{72.386} & 186.934 & 1434.554 & 0.092 & 0.778 & 0.301 & \underline{0.320} & \underline{0.298} & \underline{0.276} & -- & -- \\
\midrule
\multirow{7}{*}{\rotatebox[origin=c]{90}{\textsc{MIT-FV}}}
& GazeXplain~\cite{chen2024gazexplain} & 45.582 & 183.752 & 859.060 & 0.098 & 0.805 & 0.347 & 0.375 & 0.335 & 0.359 & -- & -- \\
& HAT~\cite{yang2024hat} & 60.118 & 221.411 & 1374.953 & 0.136 & 0.659 & 0.285 & -- & 0.305 & -- & -- & -- \\
& TPP-Gaze~\cite{damico2024tppgaze} & 43.426 & 184.633 & 824.114 & 0.101 & 0.777 & 0.306 & 0.294 & 0.305 & 0.280 & -- & -- \\
& DeepGaze III~\cite{kummerer2022deepgaze} & \textbf{41.517} & 183.371 & 801.176 & 0.101 & 0.684 & \underline{0.387} & -- & \underline{0.393} & -- & -- & -- \\
& DiffEye~\cite{kara2025diffeye} & 47.670 & \underline{177.656} & \underline{791.000} & \textbf{0.092} & \textbf{0.813} & \textbf{0.402} & \textbf{0.435} & \textbf{0.411} & \textbf{0.430} & -- & -- \\
& ScanDiff~\cite{scandiff} & 46.085 & \textbf{176.326} & 812.946 & \underline{0.093} & \underline{0.812} & 0.366 & \underline{0.387} & 0.374 & \underline{0.378} & -- & -- \\
\rowcolor{gray!10}
\cellcolor{white} & \method & \underline{41.641} & 187.029 & \textbf{744.834} & 0.105 & 0.773 & 0.324 & 0.365 & 0.350 & 0.358 & -- & -- \\
\midrule
\multirow{6}{*}{\rotatebox[origin=c]{90}{\textsc{COCO-TP}}}
& GazeFormer~\cite{mondal2023gazeformer} & \textbf{15.114} & \textbf{181.152} & \textbf{284.021} & \textbf{0.099} & \textbf{0.827} & \textbf{0.545} & \textbf{0.508} & \textbf{0.614} & \textbf{0.552} & \textbf{0.501} & \textbf{0.455} \\
& GazeXplain~\cite{chen2024gazexplain} & 15.926 & 200.329 & 307.095 & \underline{0.104} & \underline{0.807} & \underline{0.508} & \underline{0.476} & 0.591 & \underline{0.532} & 0.474 & \underline{0.439} \\
& HAT~\cite{yang2024hat} & 18.652 & 224.163 & 433.123 & 0.144 & 0.651 & 0.467 & -- & 0.565 & -- & 0.460 & -- \\
& TPP-Gaze~\cite{damico2024tppgaze} & 44.976 & 254.692 & 851.326 & 0.166 & 0.755 & 0.173 & 0.145 & 0.230 & 0.190 & 0.199 & 0.157 \\
& ScanDiff~\cite{scandiff} & 17.325 & \underline{192.211} & \underline{332.698} & 0.117 & 0.804 & 0.485 & 0.413 & 0.583 & 0.479 & 0.467 & 0.375 \\
\rowcolor{gray!10}
\cellcolor{white} & \method & \underline{15.382} & 200.597 & 359.013 & 0.137 & 0.788 & 0.472 & 0.357 & \underline{0.610} & 0.441 & \underline{0.476} & 0.334 \\
\midrule
\multirow{6}{*}{\rotatebox[origin=c]{90}{\textsc{COCO-TA}}}
& GazeFormer~\cite{mondal2023gazeformer} & 32.439 & \textbf{188.240} & 622.325 & \underline{0.100} & \underline{0.808} & 0.339 & 0.329 & 0.345 & 0.330 & 0.359 & 0.340 \\
& GazeXplain~\cite{chen2024gazexplain} & \textbf{31.986} & \underline{189.654} & \textbf{572.228} & \textbf{0.094} & \textbf{0.813} & \textbf{0.381} & \textbf{0.371} & \textbf{0.397} & \textbf{0.382} & \textbf{0.377} & \textbf{0.365} \\
& HAT~\cite{yang2024hat} & \underline{32.378} & 200.521 & \underline{630.468} & 0.121 & 0.651 & \underline{0.340} & -- & 0.360 & -- & \underline{0.363} & -- \\
& TPP-Gaze~\cite{damico2024tppgaze} & 47.572 & 200.067 & 916.626 & 0.129 & 0.777 & 0.224 & 0.203 & 0.236 & 0.209 & 0.362 & \underline{0.344} \\
& ScanDiff~\cite{scandiff} & 35.257 & 200.334 & 654.090 & 0.115 & 0.807 & \underline{0.344} & \underline{0.332} & 0.361 & \underline{0.344} & 0.273 & 0.243 \\
\rowcolor{gray!10}
\cellcolor{white} & \method & 32.894 & 202.098 & 632.080 & 0.116 & 0.788 & \underline{0.340} & 0.320 & \underline{0.363} & 0.322 & 0.360 & 0.330 \\
\bottomrule
\end{tabular}}
\end{table*}
Following DiffEye~\cite{kara2025diffeye}, the mean protocol computes a distance or similarity score between every pair of ground-truth and generated scanpaths for each input stimulus, and averages the scores across all pairs. Results are reported in \Cref{tab:all_mean}.

\clearpage
\subsection{Best-of-\texorpdfstring{$N$}{N} Protocol}
\begin{table*}[ht!]
\centering
\caption{\textbf{Best-of-$N$ metrics across free-viewing and visual-search datasets.} Datasets: COCO-FV (COCO-FreeView), MIT-FV (MIT1003), COCO-TP (COCO-Search18, Target Present), COCO-TA (COCO-Search18, Target Absent). ScanMatch (SM), Sequence Score (SS), and Semantic Sequence Score (SSS); $\dagger$ marks the duration-aware variant. \textbf{Bold}: best; \underline{underlined}: second best within each dataset. Dashes indicate methods that do not predict durations, or metrics that do not apply (SSS is undefined for free-viewing).}
\label{tab:all_best}
\setlength{\tabcolsep}{3pt}
\renewcommand{\arraystretch}{1.05}
\resizebox{\textwidth}{!}{%
\begin{tabular}{l l ccccc cccc cc}
\toprule
\textbf{Dataset} & \textbf{Method}
& LD$\downarrow$ & DFD$\downarrow$ & DTW$\downarrow$ & TDE$\downarrow$ & MM$\uparrow$
& SM$\uparrow$ & SM$^{\dagger}\uparrow$ & SS$\uparrow$ & SS$^{\dagger}\uparrow$ & SSS$\uparrow$ & SSS$^{\dagger}\uparrow$ \\
\midrule
\multirow{7}{*}{\rotatebox[origin=c]{90}{\textsc{COCO-FV}}}
& GazeXplain~\cite{chen2024gazexplain} & 97.256 & 180.455 & 1312.892 & 0.063 & 0.838 & 0.190 & 0.169 & 0.192 & 0.173 & -- & -- \\
& HAT~\cite{yang2024hat} & \underline{71.187} & 147.122 & 1233.285 & 0.080 & 0.722 & 0.389 & -- & 0.398 & -- & -- & -- \\
& TPP-Gaze~\cite{damico2024tppgaze} & 74.643 & 134.642 & 1065.083 & 0.054 & \underline{0.878} & 0.353 & 0.302 & 0.347 & 0.305 & -- & -- \\
& DeepGaze III~\cite{kummerer2022deepgaze} & 78.371 & 136.125 & \textbf{1014.752} & \underline{0.061} & 0.729 & 0.354 & -- & 0.369 & -- & -- & -- \\
& DiffEye~\cite{kara2025diffeye} & 71.682 & 142.949 & 1132.701 & 0.063 & 0.875 & 0.387 & 0.343 & 0.384 & 0.339 & -- & -- \\
& ScanDiff~\cite{scandiff} & \textbf{70.118} & \textbf{125.174} & 1111.774 & \textbf{0.060} & \textbf{0.892} & \textbf{0.445} & \textbf{0.434} & \textbf{0.445} & \textbf{0.418} & -- & -- \\
\rowcolor{gray!10}
\cellcolor{white} & \method & 71.237 & \underline{133.596} & \underline{1050.138} & 0.071 & 0.865 & \underline{0.391} & \underline{0.426} & \underline{0.400} & \underline{0.404} & -- & -- \\
\midrule
\multirow{7}{*}{\rotatebox[origin=c]{90}{\textsc{MIT-FV}}}
& GazeXplain~\cite{chen2024gazexplain} & 42.563 & 134.709 & 655.104 & 0.080 & 0.879 & 0.422 & 0.457 & 0.439 & 0.470 & -- & -- \\
& HAT~\cite{yang2024hat} & 49.778 & 158.156 & 995.366 & 0.107 & 0.719 & 0.370 & -- & 0.425 & -- & -- & -- \\
& TPP-Gaze~\cite{damico2024tppgaze} & \underline{34.951} & 117.262 & 535.849 & \textbf{0.062} & \underline{0.888} & 0.458 & 0.436 & 0.490 & 0.467 & -- & -- \\
& DeepGaze III~\cite{kummerer2022deepgaze} & 38.182 & 117.225 & 542.954 & 0.072 & 0.744 & 0.489 & -- & 0.530 & -- & -- & -- \\
& DiffEye~\cite{kara2025diffeye} & 36.852 & 119.422 & \underline{515.773} & 0.067 & \textbf{0.897} & \textbf{0.526} & \textbf{0.543} & \textbf{0.558} & \textbf{0.565} & -- & -- \\
& ScanDiff~\cite{scandiff} & 35.347 & \textbf{113.097} & 523.003 & \underline{0.064} & \textbf{0.897} & \underline{0.491} & \underline{0.505} & \underline{0.532} & 0.539 & -- & -- \\
\rowcolor{gray!10}
\cellcolor{white} & \method & \textbf{34.812} & \underline{114.222} & \textbf{485.741} & 0.065 & 0.884 & 0.471 & 0.504 & 0.520 & \underline{0.542} & -- & -- \\
\midrule
\multirow{6}{*}{\rotatebox[origin=c]{90}{\textsc{COCO-TP}}}
& GazeFormer~\cite{mondal2023gazeformer} & 15.115 & 181.157 & 284.039 & 0.099 & 0.827 & 0.544 & 0.508 & 0.614 & 0.552 & 0.501 & 0.456 \\
& GazeXplain~\cite{chen2024gazexplain} & 14.731 & 179.029 & 266.416 & 0.092 & 0.863 & 0.574 & 0.560 & 0.632 & 0.612 & 0.544 & 0.541 \\
& HAT~\cite{yang2024hat} & \textbf{11.086} & \underline{67.599} & 138.744 & 0.058 & 0.756 & \textbf{0.712} & -- & \underline{0.825} & -- & \underline{0.719} & -- \\
& TPP-Gaze~\cite{damico2024tppgaze} & 23.307 & 164.028 & 481.844 & 0.121 & 0.874 & 0.305 & 0.238 & 0.410 & 0.312 & 0.366 & 0.263 \\
& ScanDiff~\cite{scandiff} & 11.299 & 69.390 & \underline{137.772} & \underline{0.057} & \underline{0.923} & 0.689 & \underline{0.584} & 0.804 & \underline{0.666} & 0.685 & \underline{0.576} \\
\rowcolor{gray!10}
\cellcolor{white} & \method & \underline{11.104} & \textbf{60.648} & \textbf{128.501} & \textbf{0.055} & \textbf{0.925} & \underline{0.710} & \textbf{0.595} & \textbf{0.837} & \textbf{0.684} & \textbf{0.740} & \textbf{0.629} \\
\midrule
\multirow{6}{*}{\rotatebox[origin=c]{90}{\textsc{COCO-TA}}}
& GazeFormer~\cite{mondal2023gazeformer} & 32.439 & 188.240 & 622.325 & 0.100 & 0.808 & 0.339 & 0.329 & 0.345 & 0.330 & 0.359 & 0.340 \\
& GazeXplain~\cite{chen2024gazexplain} & 30.281 & 146.649 & 464.407 & 0.074 & 0.885 & 0.461 & 0.452 & 0.485 & 0.482 & 0.469 & 0.471 \\
& HAT~\cite{yang2024hat} & 27.281 & 111.527 & 395.514 & 0.069 & 0.738 & 0.491 & -- & 0.541 & -- & 0.547 & -- \\
& TPP-Gaze~\cite{damico2024tppgaze} & 31.723 & 130.847 & 573.821 & 0.091 & 0.888 & 0.359 & 0.311 & 0.407 & 0.350 & 0.444 & 0.376 \\
& ScanDiff~\cite{scandiff} & \textbf{26.136} & \textbf{103.716} & 385.415 & \textbf{0.066} & \textbf{0.911} & \textbf{0.512} & \underline{0.481} & \underline{0.546} & \underline{0.533} & \underline{0.559} & \underline{0.531} \\
\rowcolor{gray!10}
\cellcolor{white} & \method & \underline{26.738} & \underline{105.025} & \textbf{376.639} & \underline{0.068} & \underline{0.907} & \underline{0.511} & \textbf{0.484} & \textbf{0.557} & \textbf{0.541} & \textbf{0.568} & \textbf{0.542} \\
\bottomrule
\end{tabular}}
\end{table*}
Following DiffEye~\cite{kara2025diffeye}, the best-of-$N$ protocol selects, for each ground-truth scanpath, the closest match among all generated scanpaths for the same stimulus, using the minimum score for distance metrics and the maximum score for similarity metrics. We then average these best scores across all ground-truth scanpaths. Results are reported in \Cref{tab:all_best}.

\clearpage
\subsection{KLD Protocol}
\begin{table*}[ht!]
\centering
\caption{\textbf{KLD metrics across free-viewing and visual-search datasets.} Datasets: COCO-FV (COCO-FreeView), MIT-FV (MIT1003), COCO-TP (COCO-Search18, Target Present), COCO-TA (COCO-Search18, Target Absent). ScanMatch (SM), Sequence Score (SS), and Semantic Sequence Score (SSS); $\dagger$ marks the duration-aware variant. \textbf{Bold}: best; \underline{underlined}: second best within each dataset. Dashes indicate methods that do not predict durations, or metrics that do not apply (SSS is undefined for free-viewing).}
\label{tab:all_kld}
\setlength{\tabcolsep}{3pt}
\renewcommand{\arraystretch}{1.05}
\resizebox{\textwidth}{!}{%
\begin{tabular}{l l ccccc cccc cc}
\toprule
\textbf{Dataset} & \textbf{Method}
& LD$\downarrow$ & DFD$\downarrow$ & DTW$\downarrow$ & TDE$\downarrow$ & MM$\downarrow$
& SM$\downarrow$ & SM$^{\dagger}\downarrow$ & SS$\downarrow$ & SS$^{\dagger}\downarrow$ & SSS$\downarrow$ & SSS$^{\dagger}\downarrow$ \\
\midrule
\multirow{7}{*}{\rotatebox[origin=c]{90}{\textsc{COCO-FV}}}
& GazeXplain~\cite{chen2024gazexplain} & 0.630 & 0.116 & \underline{0.026} & 0.322 & 0.369 & 3.412 & 4.907 & 5.640 & 2.407 & -- & -- \\
& HAT~\cite{yang2024hat} & 0.115 & 0.191 & 0.149 & 1.002 & 4.218 & 0.559 & -- & 0.344 & -- & -- & -- \\
& TPP-Gaze~\cite{damico2024tppgaze} & 0.163 & \underline{0.010} & \underline{0.026} & \underline{0.037} & 0.077 & 1.001 & 1.914 & 0.774 & 0.619 & -- & -- \\
& DeepGaze III~\cite{kummerer2022deepgaze} & 0.152 & 0.011 & 0.077 & \textbf{0.036} & 3.855 & 0.682 & -- & 1.266 & -- & -- & -- \\
& DiffEye~\cite{kara2025diffeye} & \textbf{0.098} & \underline{0.010} & 0.043 & 0.038 & \textbf{0.022} & \underline{0.230} & 0.522 & \underline{0.248} & \underline{0.120} & -- & -- \\
& ScanDiff~\cite{scandiff} & \underline{0.100} & 0.036 & \textbf{0.011} & 0.045 & \underline{0.070} & \textbf{0.022} & \textbf{0.042} & \textbf{0.036} & \textbf{0.012} & -- & -- \\
\rowcolor{gray!10}
\cellcolor{white} & \method & 0.314 & \textbf{0.005} & 0.053 & 0.062 & 0.237 & 0.676 & \underline{0.103} & 0.382 & \underline{0.102} & -- & -- \\
\midrule
\multirow{7}{*}{\rotatebox[origin=c]{90}{\textsc{MIT-FV}}}
& GazeXplain~\cite{chen2024gazexplain} & 1.137 & 0.041 & 0.075 & 0.041 & 0.086 & 0.202 & 0.071 & 2.197 & 0.072 & -- & -- \\
& HAT~\cite{yang2024hat} & 4.237 & 1.554 & 3.986 & 2.985 & 4.572 & 1.937 & -- & 1.743 & -- & -- & -- \\
& TPP-Gaze~\cite{damico2024tppgaze} & \textbf{0.020} & 0.015 & 0.027 & \textbf{0.012} & 0.074 & 0.342 & 0.575 & 0.289 & 0.428 & -- & -- \\
& DeepGaze III~\cite{kummerer2022deepgaze} & 0.741 & 0.031 & 0.113 & 0.075 & 3.721 & 0.106 & -- & 1.035 & -- & -- & -- \\
& DiffEye~\cite{kara2025diffeye} & 0.080 & \textbf{0.011} & \underline{0.018} & \underline{0.021} & \textbf{0.024} & \textbf{0.014} & \textbf{0.018} & 0.071 & \textbf{0.018} & -- & -- \\
& ScanDiff~\cite{scandiff} & 0.041 & \underline{0.014} & 0.019 & 0.037 & \underline{0.044} & \underline{0.073} & \underline{0.045} & \textbf{0.055} & \underline{0.032} & -- & -- \\
\rowcolor{gray!10}
\cellcolor{white} & \method & \underline{0.029} & 0.021 & \textbf{0.013} & 0.038 & 0.095 & 0.164 & 0.056 & \underline{0.142} & 0.038 & -- & -- \\
\midrule
\multirow{6}{*}{\rotatebox[origin=c]{90}{\textsc{COCO-TP}}}
& GazeFormer~\cite{mondal2023gazeformer} & 0.169 & \textbf{0.039} & 0.097 & 0.164 & 0.111 & \textbf{0.107} & \underline{0.063} & 0.109 & 0.104 & \underline{0.072} & 0.127 \\
& GazeXplain~\cite{chen2024gazexplain} & 0.324 & 0.061 & 0.135 & 0.238 & 0.119 & 0.149 & \textbf{0.018} & 0.176 & 0.061 & 0.130 & 0.145 \\
& HAT~\cite{yang2024hat} & 0.147 & 0.149 & 0.183 & 0.176 & 3.835 & 0.170 & -- & 0.151 & -- & 0.140 & -- \\
& TPP-Gaze~\cite{damico2024tppgaze} & 2.893 & 2.265 & 3.736 & 3.209 & 0.715 & 5.206 & 5.886 & 3.170 & 3.137 & 2.229 & 1.751 \\
& ScanDiff~\cite{scandiff} & \underline{0.090} & \underline{0.047} & \underline{0.078} & \underline{0.095} & \underline{0.051} & \underline{0.109} & 0.065 & \underline{0.085} & \textbf{0.022} & 0.085 & \textbf{0.076} \\
\rowcolor{gray!10}
\cellcolor{white} & \method & \textbf{0.028} & 0.068 & \textbf{0.066} & \textbf{0.081} & \textbf{0.042} & 0.152 & 0.206 & \textbf{0.045} & \underline{0.048} & \textbf{0.054} & \underline{0.090} \\
\midrule
\multirow{6}{*}{\rotatebox[origin=c]{90}{\textsc{COCO-TA}}}
& GazeFormer~\cite{mondal2023gazeformer} & 0.118 & 0.085 & 0.069 & 0.109 & 0.093 & 0.038 & 0.030 & 0.235 & 0.101 & 0.152 & 0.086 \\
& GazeXplain~\cite{chen2024gazexplain} & 0.112 & 0.027 & 0.066 & 0.109 & 0.044 & 0.033 & 0.041 & 0.162 & 0.044 & 0.188 & 0.041 \\
& HAT~\cite{yang2024hat} & 0.062 & 0.024 & 0.036 & 0.044 & 3.799 & 0.042 & -- & 0.106 & -- & 0.070 & -- \\
& TPP-Gaze~\cite{damico2024tppgaze} & 0.665 & 0.182 & 0.476 & 0.266 & 0.129 & 0.752 & 0.917 & 0.540 & 0.628 & 0.281 & 0.227 \\
& ScanDiff~\cite{scandiff} & \textbf{0.022} & \underline{0.012} & \textbf{0.014} & \textbf{0.007} & \underline{0.020} & \textbf{0.012} & \textbf{0.008} & \textbf{0.031} & \textbf{0.008} & \textbf{0.020} & \underline{0.008} \\
\rowcolor{gray!10}
\cellcolor{white} & \method & \underline{0.039} & \textbf{0.007} & \underline{0.024} & \underline{0.016} & \textbf{0.018} & \underline{0.018} & \underline{0.014} & \underline{0.045} & \underline{0.011} & \underline{0.039} & \textbf{0.006} \\
\bottomrule
\end{tabular}}
\end{table*}
Following ScanDiff~\cite{scandiff} and TPP-Gaze~\cite{damico2024tppgaze}, the KLD protocol pools human-consistency scores from ground-truth versus ground-truth scanpath pairs and model-alignment scores from ground-truth versus generated pairs across the evaluation set, converts each pooled score set into a normalized histogram, and reports the KL divergence between the human and model score distributions. Results are reported in \Cref{tab:all_kld}.


\clearpage

\section{Background on Model Architecture, Training, and Sampling}
\label{sec:scandiff_preliminaries}
Our model architecture primarily follows ScanDiff~\cite{scandiff}, a conditional diffusion model for scanpath generation. Given an image stimulus $I\in\mathbb{R}^{H\times W\times 3}$ and a viewing task $c$, the goal is to generate a scanpath $S=\{f_m\}_{m=1}^{M}$, where each fixation $f_m=(r_m,\Delta t_m^{\mathrm{sp}})$ contains a 2D gaze location $r_m=(x_m^{\mathrm{sp}},y_m^{\mathrm{sp}})$ and a fixation duration $\Delta t_m^{\mathrm{sp}}$. Rather than diffusing directly in the raw scanpath space, ScanDiff first maps the scanpath into a latent token space through a learnable projection $g_{\theta}:\mathbb{R}^{M\times 3}\rightarrow\mathbb{R}^{M\times d}$, producing $z_0=g_{\theta}(S)$. The forward process gradually corrupts $z_0$ into noisy latents $z_k$ with a Gaussian diffusion process:
\[
q(z_k\mid z_{k-1})=\mathcal{N}\left(z_k;\sqrt{1-\beta_k}z_{k-1},\beta_k I\right).
\]
The conditional reverse process is then defined as
\[
p_{\theta}(z_{0:K}\mid I,c)=p(z_K)\prod_{k=1}^{K}p_{\theta}(z_{k-1}\mid z_k,I,c).
\]
The denoising network $\phi_{\theta}$ is implemented as an encoder-only Transformer with an additional cross-attention layer inserted between self-attention and feed-forward layers. Self-attention models dependencies among scanpath tokens, while cross-attention conditions the denoising process on visual and task information. Specifically, ScanDiff extracts dense image features $v(I)\in\mathbb{R}^{h\times w\times d_v}$ using a DINOv2 \cite{oquab2023dinov2} visual encoder and encodes the viewing task with a CLIP \cite{radford2021learningclip} text encoder $\psi(c)\in\mathbb{R}^{d_t}$. The visual and textual features are projected into a shared $d$-dimensional space, concatenated along the channel dimension, and linearly projected to obtain the joint multimodal feature $V_{\mathrm{joint}}\in\mathbb{R}^{hw\times d}$. The noisy latent sequence $z_k$ is augmented with positional and diffusion timestep embeddings, and the denoised latent prediction is obtained as $\hat{z}_0=\phi_{\theta}(z_k,V_{\mathrm{joint}})$. A feed-forward decoder $\gamma_{\theta}$ reconstructs the scanpath $\hat{S}=\gamma_{\theta}(\hat{z}_0)$, while a length prediction head $l_{\theta}:\mathbb{R}^{M\times d}\rightarrow\mathbb{R}^{M}$ predicts token validity for variable-length generation.
ScanDiff is trained with a combination of four losses:
\[
\mathcal{L}
=
\mathcal{L}_{\mathrm{diff}}
+
\mathcal{L}_{\mathrm{recon}}
+
\mathcal{L}_{\mathrm{val}}
+
\mathcal{L}_{\mathrm{reg}}.
\]
The latent diffusion loss encourages the denoised prediction to recover the clean latent sequence:
\[
\mathcal{L}_{\mathrm{diff}}
=
\sum_{k=1}^{K}
\left\|
z_0-\phi_{\theta}(z_k,V_{\mathrm{joint}})
\right\|_2^2 .
\]
The reconstruction loss supervises the decoded scanpath with an $\ell_1$ loss over fixation coordinates and durations:
\[
\mathcal{L}_{\mathrm{recon}}
=
\frac{1}{M}
\sum_{m=1}^{M}
\left(
|x_m^{\mathrm{sp}}-\hat{x}_m^{\mathrm{sp}}|
+
|y_m^{\mathrm{sp}}-\hat{y}_m^{\mathrm{sp}}|
+
|\Delta t_m^{\mathrm{sp}}-\Delta\hat{t}_m^{\mathrm{sp}}|
\right),
\]
where padded tokens are masked out. The validity loss $\mathcal{L}_{\mathrm{val}}$ is a binary cross-entropy loss between the predicted token-validity probabilities and the ground-truth validity labels, and $\mathcal{L}_{\mathrm{reg}}=\|\mu(z_K)\|_2^2$ regularizes the terminal diffusion state so that it matches a standard Gaussian prior.
At inference time, ScanDiff samples an initial latent sequence $z_K\sim\mathcal{N}(0,I)$ and iteratively applies the learned reverse process conditioned on $V_{\mathrm{joint}}$ until obtaining $\hat{z}_0$. The decoded output $\hat{S}=\gamma_{\theta}(\hat{z}_0)$ gives the predicted fixation coordinates and durations, while the length head determines the valid prefix of the sequence. Multiple diverse scanpaths for the same image and task are produced by sampling different initial noise sequences $z_K$.
\clearpage

\section{Implementation Details}
\label{sec:implementation_details}
\subsection{Training Details}
We implement all models in PyTorch and train them using AdamW with a learning rate of $1\times10^{-4}$ and a weight decay of $1\times10^{-2}$.
All models are trained for 300 epochs on a single NVIDIA GH200 GPU using internal clusters. The training time of each model is about 1 day.
Unless otherwise specified, other training configurations and model architectures, such as length prediction and image and text conditioning, follow the official implementation of ScanDiff \cite{scandiff}.
We also follow ScanDiff to train separate models for different datasets to account for their different task conditions and data distributions. Empirically, the best fixation-index injection point varies by dataset, with latent-space augmentation performing slightly better on MIT1003.

\clearpage
\subsection{Dataset-Specific Implementation Details}
\label{sec:mit_details}
\begin{table*}[ht!]
\centering
\caption{\textbf{Ablation study on concatenation location (MIT1003).} We compare appending the fixation index at the raw data-channel level versus the latent-channel level. Metrics follow the main paper CRPS protocol; $\dagger$ marks duration-aware variants.}
\vspace{8pt}
\label{tab:ablation_mit}
\small
\resizebox{\textwidth}{!}{%
\begin{tabular}{@{} l *{9}{c} @{}}
\toprule
\textbf{Concat Location}
& LD$\downarrow$ & DFD$\downarrow$ & DTW$\downarrow$ & TDE$\downarrow$ & MM$\uparrow$
& SM$\uparrow$ & SM$^{\dagger}\uparrow$ & SS$\uparrow$ & SS$^{\dagger}\uparrow$ \\
\midrule
Data channel & 26.706 & 108.258 & 494.392 & 0.062 & 0.371 & 0.104 & 0.128 & 0.281 & 0.127 \\
Latent space & \textbf{24.961} & \textbf{100.903} & \textbf{456.985} & \textbf{0.058} & \textbf{0.380} & \textbf{0.129} & \textbf{0.145} & \textbf{0.321} & \textbf{0.146} \\
\bottomrule
\end{tabular}}
\end{table*}
Through empirical experiments, we found that injecting the fixation index in the latent space performs better on MIT1003. This may be due to the noisier trajectory annotations in MIT1003, where some raw trajectories move out of frame between fixations, making coordinate-level trajectory-scanpath alignment less reliable. Latent-space injection can partially reduce the effect of this misalignment. Specifically, instead of concatenating the fixation index as an additional raw trajectory channel, we first project the trajectory into the latent sequence of size $T \times 256$ and then concatenate the fixation index in the latent space, resulting in a $T \times 257$ latent representation. During sampling, we initialize Gaussian noise with the same latent dimensionality and discard the additional fixation-index dimension before passing the denoised latent sequence to the decoder. All other settings follow the experiments on the remaining datasets. We show the numbers in \Cref{tab:ablation_mit}.

\clearpage
\section{Statistical Analysis of Generated Scanpaths}
\label{sec:scanpath_stats}
\begin{figure*}[ht!]
    \centering
    \includegraphics[width=1\textwidth]{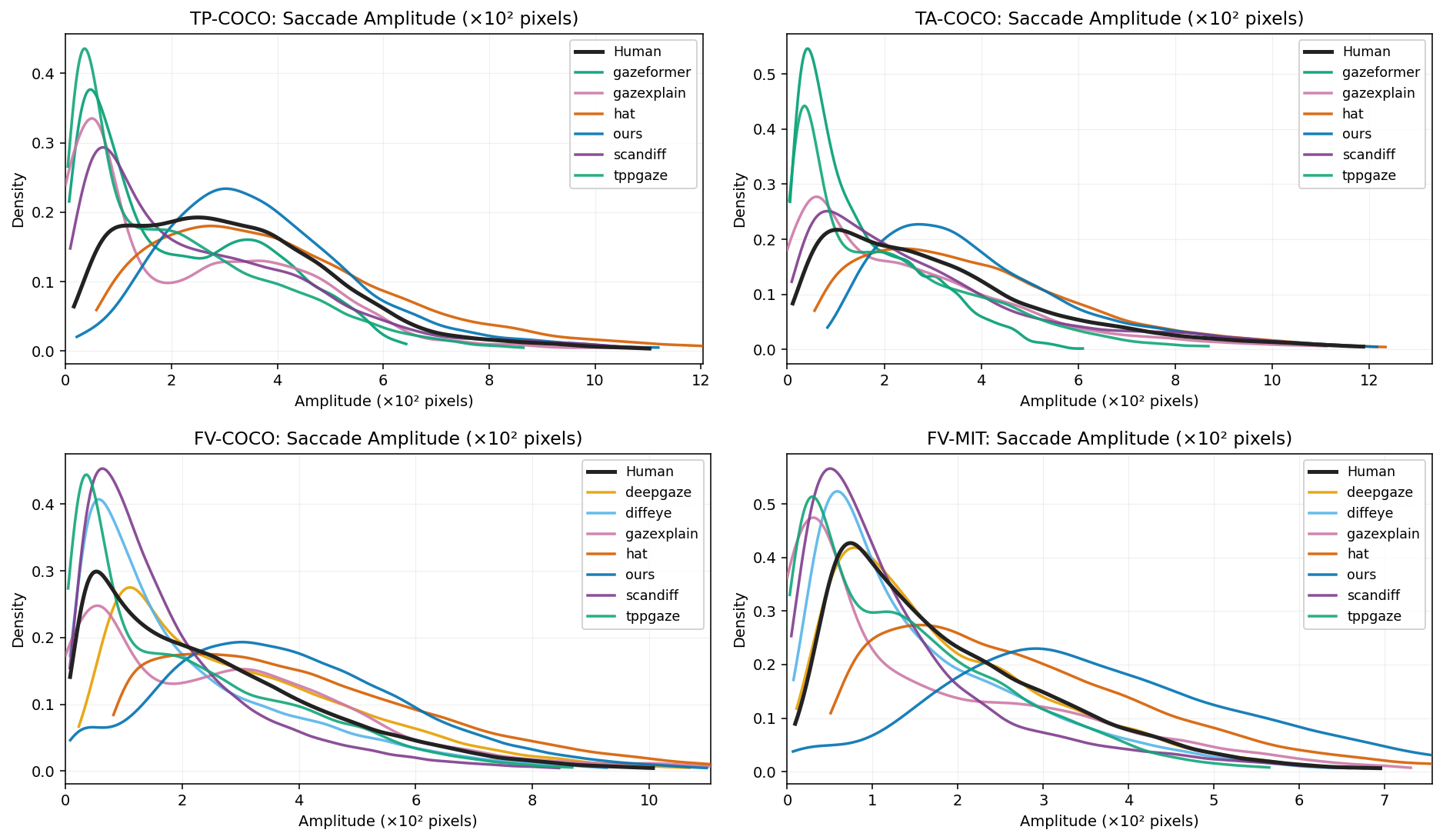} 
    \caption{\textbf{Distributions of saccade amplitude for different datasets.} The horizontal axis represents the amplitude (length) of saccades in the unit of $10^2$ pixels. The vertical axis represents the probability density.}
    \label{fig:distribution_amplitude}
\end{figure*}

\begin{figure*}[ht!]
    \centering
    \includegraphics[width=1\textwidth]{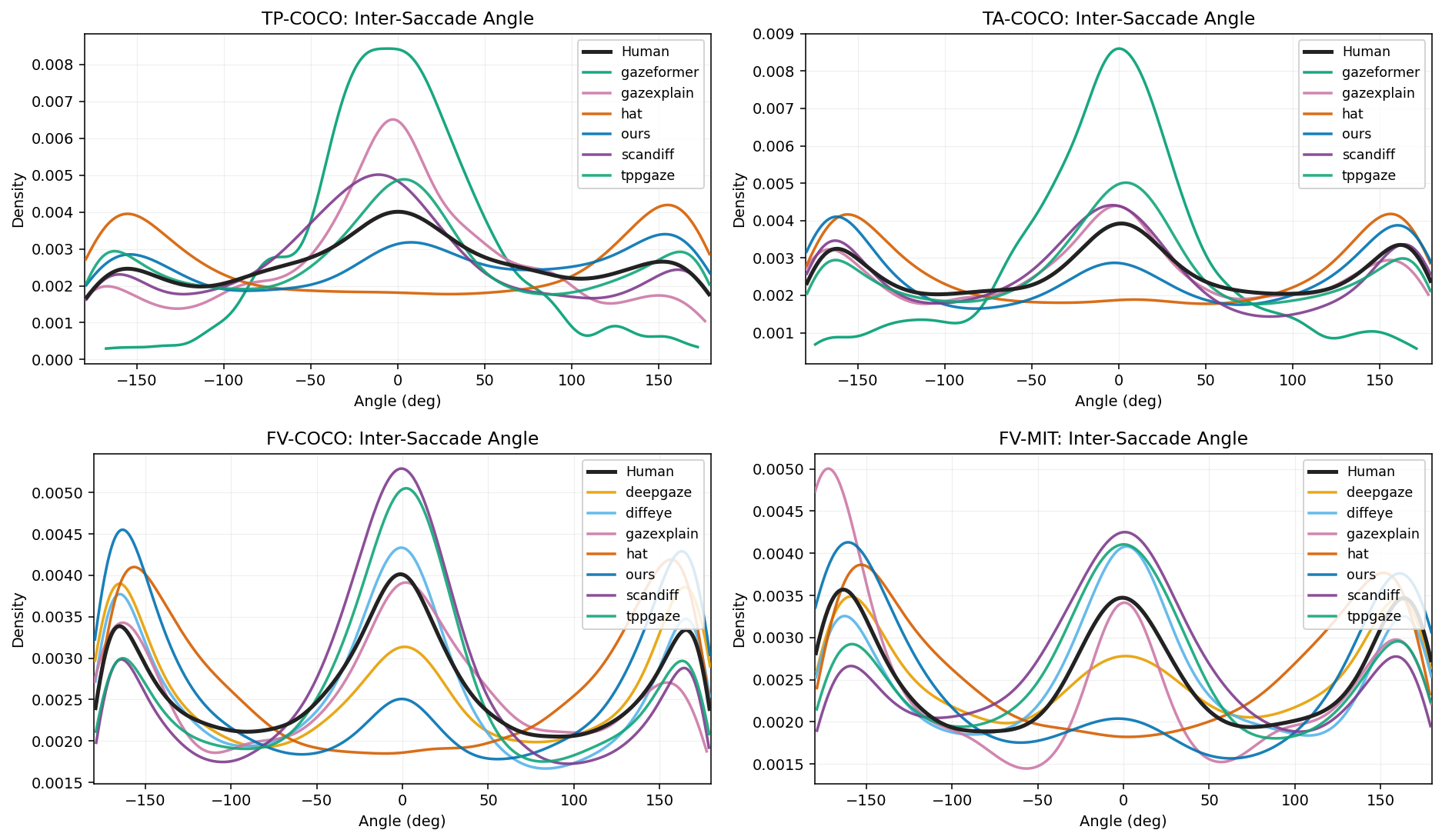} 
    \caption{\textbf{Distributions of inter-saccade angle for different datasets.} The horizontal axis represents the degree of angles between consecutive saccades. The vertical axis represents the probability density.}
    \label{fig:distribution_angle}
\end{figure*}

\begin{figure*}[ht!]
    \centering
    \includegraphics[width=1\textwidth]{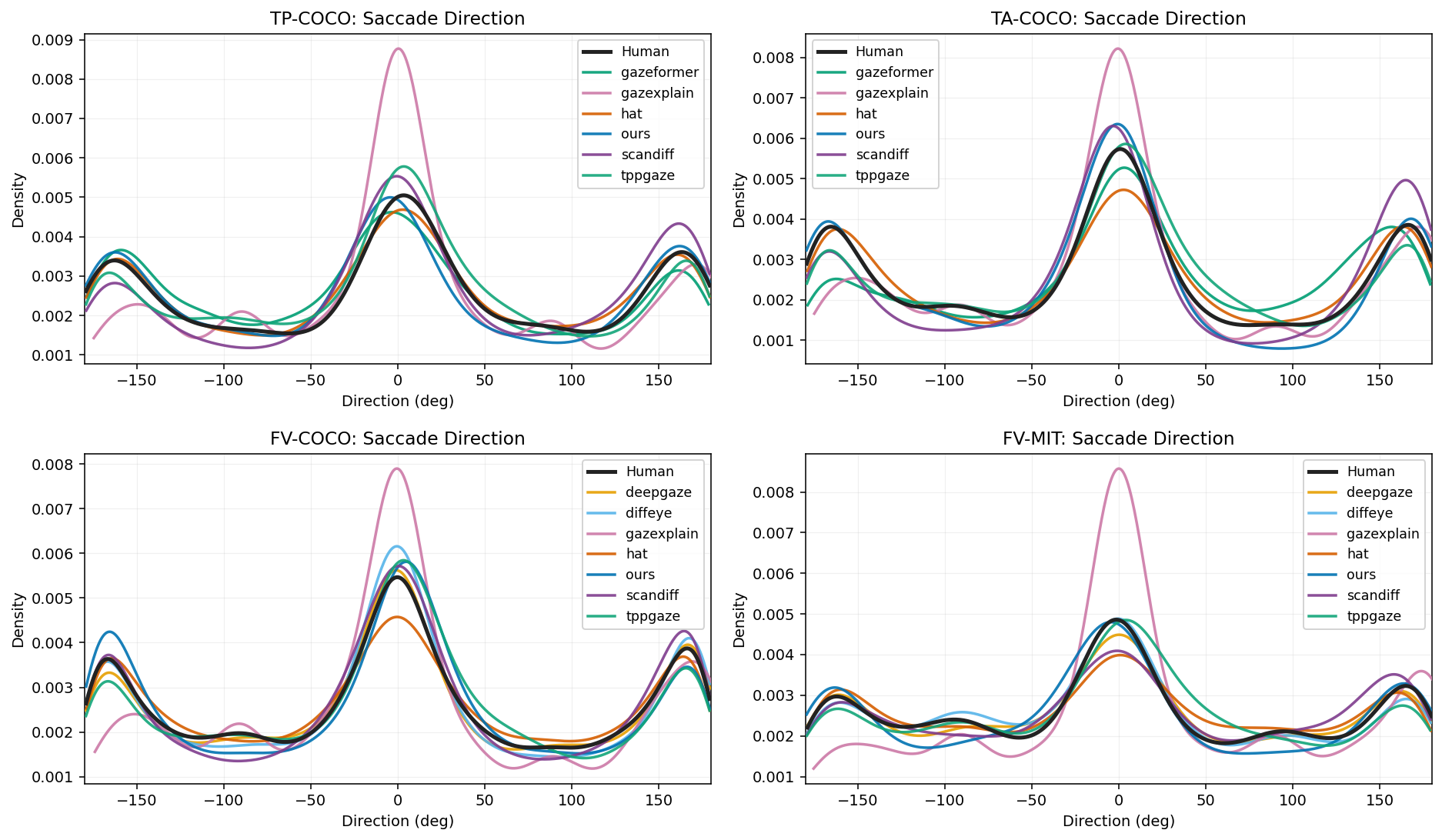} 
    \caption{\textbf{Distributions of saccade direction for different datasets.} The horizontal axis represents the direction of the saccades in degrees. The vertical axis represents the probability density.}
    \label{fig:distribution_direction}
\end{figure*}
We analyze the statistical distributions of real and generated scanpaths in terms of saccade amplitude, saccade direction, and inter-saccade angle. The corresponding plots are shown in \Cref{fig:distribution_amplitude,fig:distribution_angle,fig:distribution_direction}. The amplitude distributions indicate that human saccades typically follow a long-tailed pattern, with a peak around 100 pixels. The generated scanpaths generally follow a similar amplitude distribution. For inter-saccade angles, we observe peaks near 0 and 180 degrees. This suggests that gaze transitions often either continue in the previous direction or reverse to the opposite direction, indicating that consecutive eye movements are frequently colinear. For saccade direction, the overall distribution shapes are also similar between real and generated scanpaths. Both show a stronger horizontal bias than other directions. Overall, these results suggest that the generated scanpaths capture the key statistical properties of human eye movements.

\clearpage
\section{Additional Qualitative Results}
\label{sec:additional_qualitative}

\subsection{Additional Scanpath Results}
\label{sec:additional_scanpath_results}
\begin{figure}[ht!]
    \centering
    \includegraphics[width=\textwidth]{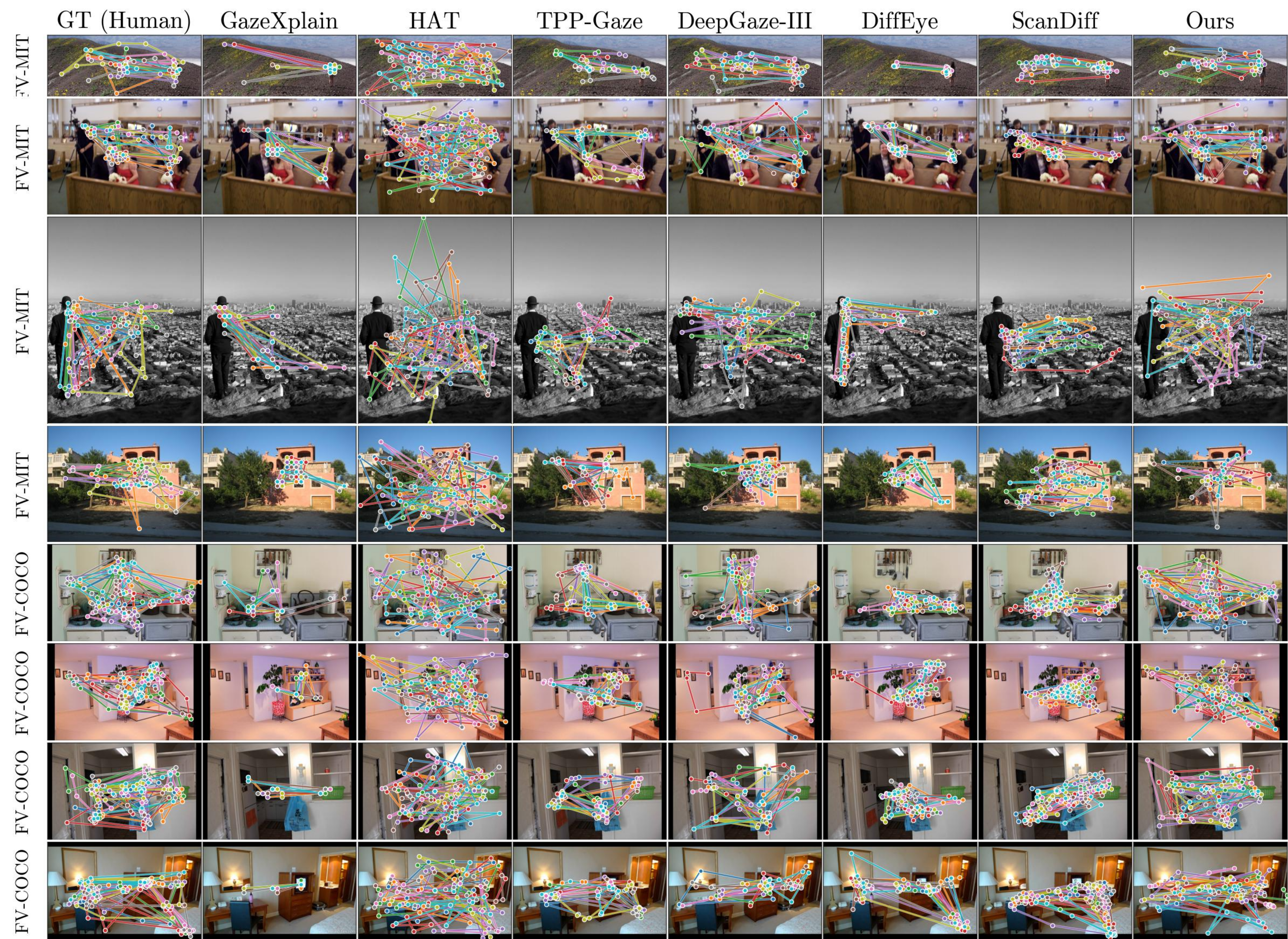}
    \caption{\textbf{Additional scanpath comparisons on free-viewing datasets.} Generated scanpaths from our method and baselines on COCO-FreeView and MIT1003.}
    \label{fig:appendix_freeview}
\end{figure}

\begin{figure}[ht!]
    \centering
    \includegraphics[width=\textwidth]{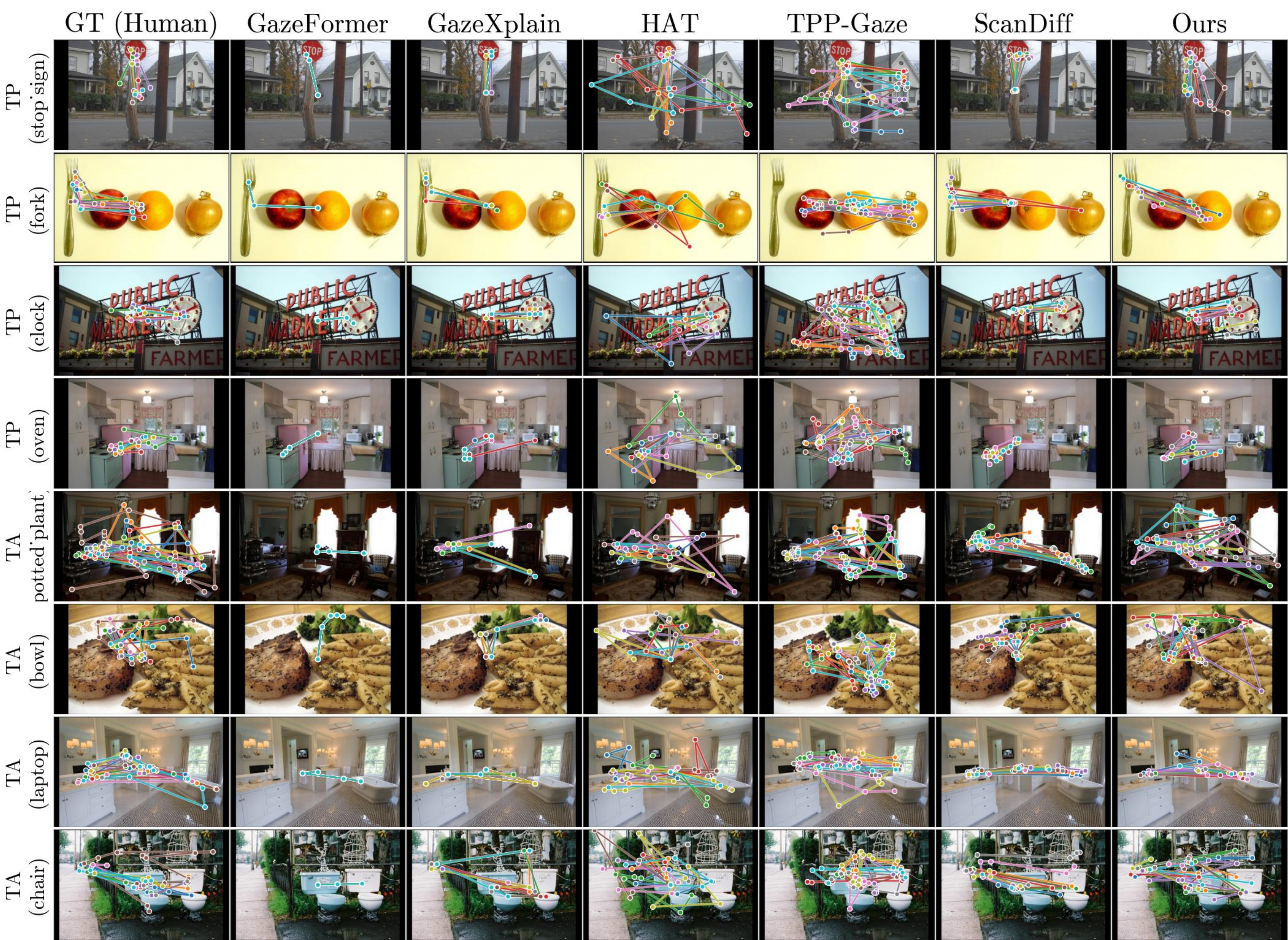}
    \caption{\textbf{Additional scanpath comparisons on visual-search datasets.} Generated scanpaths from our method and baselines on COCO-Search18 (target-present and target-absent).}
    \label{fig:appendix_search}
\end{figure}

We provide additional qualitative comparisons of generated scanpaths against baselines on the free-viewing benchmarks (COCO-FreeView and MIT1003) in \Cref{fig:appendix_freeview} and on the visual-search benchmark (COCO-Search18, target-present and target-absent) in \Cref{fig:appendix_search}. Across both settings, our model produces scanpaths that better align with the spatial structure and exploration patterns of the human ground truth, while baselines tend to either collapse to low-diversity outputs or drift away from the salient regions covered by human viewers.

\subsection{Additional Trajectory Results}
\label{sec:additional_trajectory_results}
\begin{figure}[ht!]
    \centering
    \includegraphics[width=\textwidth]{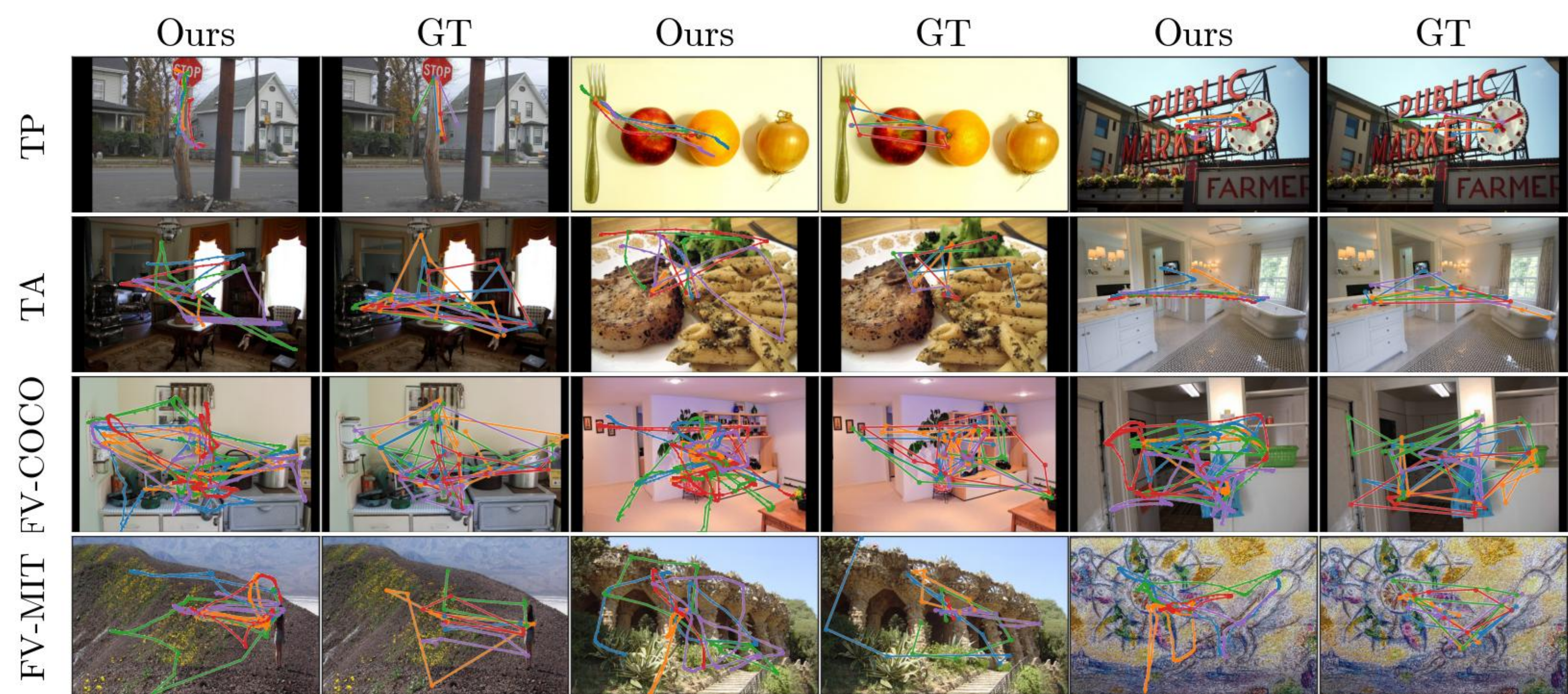}
    \caption{\textbf{Additional trajectory comparisons.} Generated continuous trajectories from our method and DiffEye~\cite{kara2025diffeye} across free-viewing and visual-search datasets.}
    \label{fig:appendix_trajectories}
\end{figure}

We further compare generated continuous trajectories against the only existing continuous-trajectory baseline, DiffEye~\cite{kara2025diffeye}, in \Cref{fig:appendix_trajectories}. Our trajectories more faithfully reproduce the spatial coverage and density patterns of the human ground truth, capturing both the fixation clusters and the saccadic transitions between them. In contrast, DiffEye tends to produce smoother, less structured paths that under-represent the discrete fixation organization characteristic of human gaze.

\clearpage
\section{Limitations and Future Work}
\label{sec:limitations}
\begin{figure*}[ht!]
    \centering
    \includegraphics[width=1\textwidth]{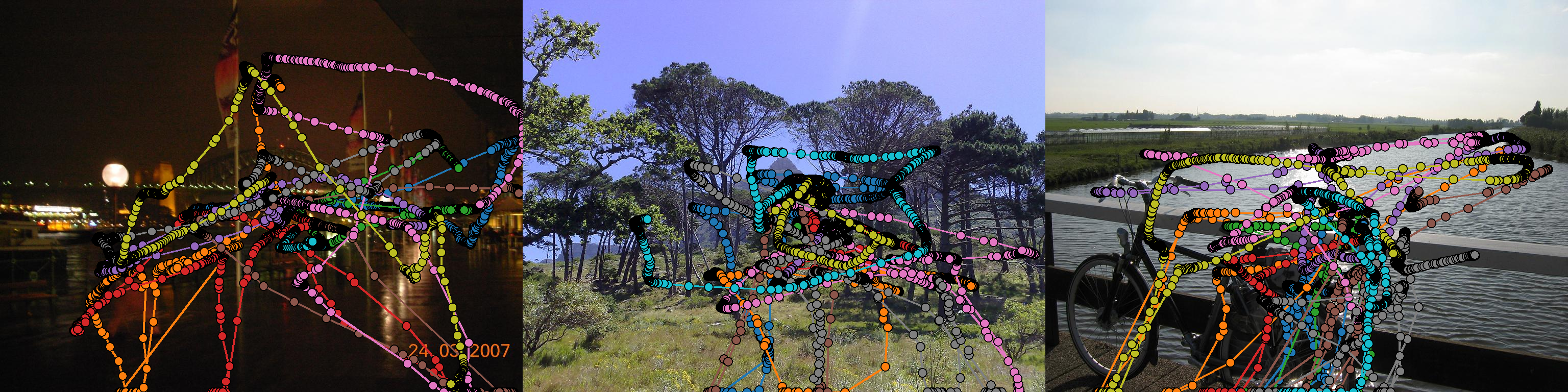} 
    \caption{\textbf{Examples of failure cases caused by generated trajectories going out of frame.} Some generated trajectories tend to go outside the image frame, which can lead to inaccurate scanpaths after trajectory-to-scanpath conversion.}
    \label{fig:limitations}
\end{figure*}
While joint trajectory-scanpath modeling improves generation quality, it also introduces higher computational cost. Compared with scanpaths, raw trajectories are typically hundreds to thousands of times longer, which increases memory usage, training time, and sampling cost. Although our formulation only requires a simple channel expansion, the long temporal horizon of trajectory data remains a practical bottleneck. Future work can explore more efficient architectures for joint trajectory-scanpath modeling, such as temporal compression, hierarchical diffusion, or coarse-to-fine generation. We show examples of failed trajectory generations in \Cref{fig:limitations}.
Another limitation comes from the stochastic and noisy nature of raw trajectory data. Unlike scanpaths, trajectories may contain abnormal patterns such as blinks, missing samples, or out-of-frame gaze points. These artifacts can affect training and may lead the model to generate unrealistic or out-of-bound trajectories. Future work can investigate stronger trajectory regularization, automatic detection of abnormal gaze patterns, and preprocessing strategies that better align trajectory signals with their corresponding scanpath structure.

\clearpage
\section{Broader Impacts}
\label{sec:broader_impacts}
Modeling human gaze has several positive societal applications. More accurate generative gaze models can support research in cognitive science, psychology, human-computer interaction, accessibility, virtual and augmented reality, and visual interface design. In clinical settings, gaze modeling may help study atypical visual attention patterns and support screening or diagnosis of developmental and mental health conditions, such as autism spectrum disorder or attention-related disorders, when used together with appropriate clinical expertise. Synthetic gaze generation may also reduce the cost of collecting large-scale eye-tracking data and enable safer evaluation of gaze-aware systems without exposing private user recordings.
At the same time, gaze data is sensitive because it can reveal information about attention, intent, cognitive state, preference, and potentially health-related traits. Misuse of gaze modeling could enable invasive user profiling, manipulative interface or advertisement design, surveillance, or attention monitoring without informed consent. The technology may also be applied in high-stakes or ethically sensitive domains, including military training or assessment systems. These risks highlight the need for careful data governance, informed consent, privacy-preserving preprocessing, and restrictions on deployment in settings where gaze data could be used to infer sensitive personal attributes or manipulate user behavior. Our work focuses on improving scientific modeling and evaluation of gaze generation, and we encourage future uses to follow ethical review, transparency, and privacy safeguards.

\clearpage
\section{Asset Licenses}
\label{sec:licenses}
Our code is mainly developed based on ScanDiff~\cite{scandiff} and DiffEye~\cite{kara2025diffeye}. ScanDiff is open-source on GitHub, but no explicit license is specified. DiffEye is open-source on GitHub under the MIT License.
The datasets used in this work, including COCO-Search18~\cite{chen2021cocosearch18}, COCO-FreeView~\cite{chen2021predictingcocofreeview}, and MIT1003~\cite{Judd_2009mit1003}, are distributed under the MIT License. We agree to follow the terms, conditions, and licenses of all codebases and datasets used in this work.



\end{document}